\documentclass[10pt,twocolumn,letterpaper]{article}

\usepackage{cvpr}              
\usepackage{soul}
\definecolor{turquoise}{cmyk}{0.65,0,0.1,0.3}
\definecolor{purple}{rgb}{0.65,0,0.65}
\definecolor{dark_green}{rgb}{0, 0.5, 0}
\definecolor{orange}{rgb}{0.8, 0.6, 0.2}
\definecolor{red}{rgb}{0.8, 0.2, 0.2}
\definecolor{darkred}{rgb}{0.6, 0.1, 0.05}
\definecolor{blueish}{rgb}{0.0, 0.3, .6}
\definecolor{light_gray}{rgb}{0.7, 0.7, .7}
\definecolor{pink}{rgb}{1, 0, 1}
\definecolor{greyblue}{rgb}{0.25, 0.25, 1}

\definecolor{blueish}{rgb}{0.0, 0.3, .6}

\usepackage{mathtools}
\usepackage{xspace}
\usepackage{bm}

\def\argmax{\mathop{\rm arg\,max}\limits}

\newcommand{\ignore}[1]{}

\makeatletter
\DeclareRobustCommand\onedot{\futurelet\@let@token\@onedot}
\def\@onedot{\ifx\@let@token.\else.\null\fi\xspace}
\def\eg{{e.g}\onedot}

\def\ie{{i.e}\onedot}

\def\etal{{et al}\onedot}
\makeatother

\input{preamble}
\usepackage[accsupp]{axessibility} 

\definecolor{cvprblue}{rgb}{0.21,0.49,0.74}
\usepackage[pagebackref,breaklinks,colorlinks,allcolors=cvprblue]{hyperref}

\def\paperID{4876} 
\def\confName{CVPR}
\def\confYear{2025}

\title{PoseBH: Prototypical Multi-Dataset Training Beyond Human Pose Estimation}

\author{
Uyoung~Jeong${}^{1}$ \quad
Jonathan~Freer${}^{2}$ \quad
Seungryul~Baek${}^{1}$ \quad
Hyung~Jin~Chang${}^{2}$ \quad
Kwang~In~Kim${}^{3}$ \vspace{0.2cm}\\
${}^{1}$UNIST \quad ${}^{2}$University of Birmingham \quad  ${}^{3}$POSTECH \\
}

\begin{document}
\maketitle

\begin{abstract}
We study multi-dataset training (MDT) for pose estimation, where skeletal heterogeneity presents a unique challenge that existing methods have yet to address. In traditional domains, \eg regression and classification, MDT typically relies on dataset merging or multi-head supervision. However, the diversity of skeleton types and limited cross-dataset supervision complicate integration in pose estimation. To address these challenges, we introduce PoseBH, a new MDT framework that tackles keypoint heterogeneity and limited supervision through two key techniques. First, we propose nonparametric keypoint prototypes that learn within a unified embedding space, enabling seamless integration across skeleton types. Second, we develop a cross-type self-supervision mechanism that aligns keypoint predictions with keypoint embedding prototypes, providing supervision without relying on teacher-student models or additional augmentations. PoseBH substantially improves generalization across whole-body and animal pose datasets, including COCO-WholeBody, AP-10K, and APT-36K, while preserving performance on standard human pose benchmarks (COCO, MPII, and AIC). Furthermore, our learned keypoint embeddings transfer effectively to hand shape estimation (InterHand2.6M) and human body shape estimation (3DPW). The code for PoseBH is available at: \url{https://github.com/uyoung-jeong/PoseBH}.
\end{abstract}

\section{Introduction}
\label{sec:intro}
The demand for human pose estimators has grown substantially in recent years, driving their adoption across diverse applications, including 3D avatar generation~\cite{jiang2022neuman,liu2024humangaussian,moon2024exavatar}, motion generation~\cite{rempe2021humor,kapon2024mas,lin2023motionx}, human-robot interaction~\cite{vendrow2023jrdb,Le_2024_CVPR,yu2024human}, safety monitoring~\cite{cormier2022we,solbach2017vision,lamas2022human}, and VR/AR tracking~\cite{tome2019xr,guzov2021human,wang2023scene}. These domains pose distinct operational challenges, often necessitating significant domain adaptation for robust pose estimation. To address this, we aim to train a human pose estimator that generalizes effectively across multiple datasets.

\begin{figure}[t]
\centering
{
\footnotesize
\setlength\tabcolsep{1pt}
\begin{tabular}{cccc}
\includegraphics[width=0.24\linewidth]{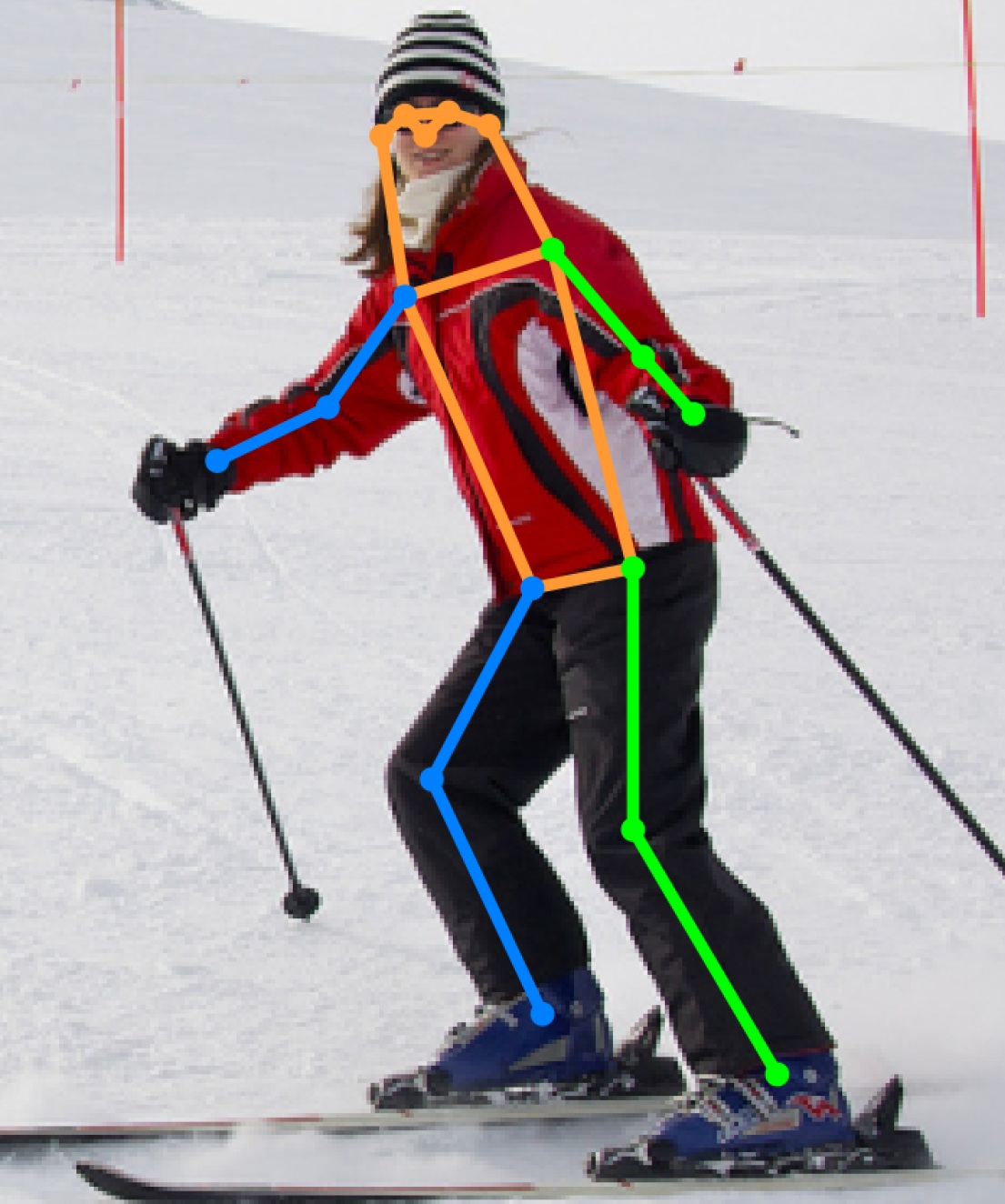} &
\includegraphics[width=0.24\linewidth]{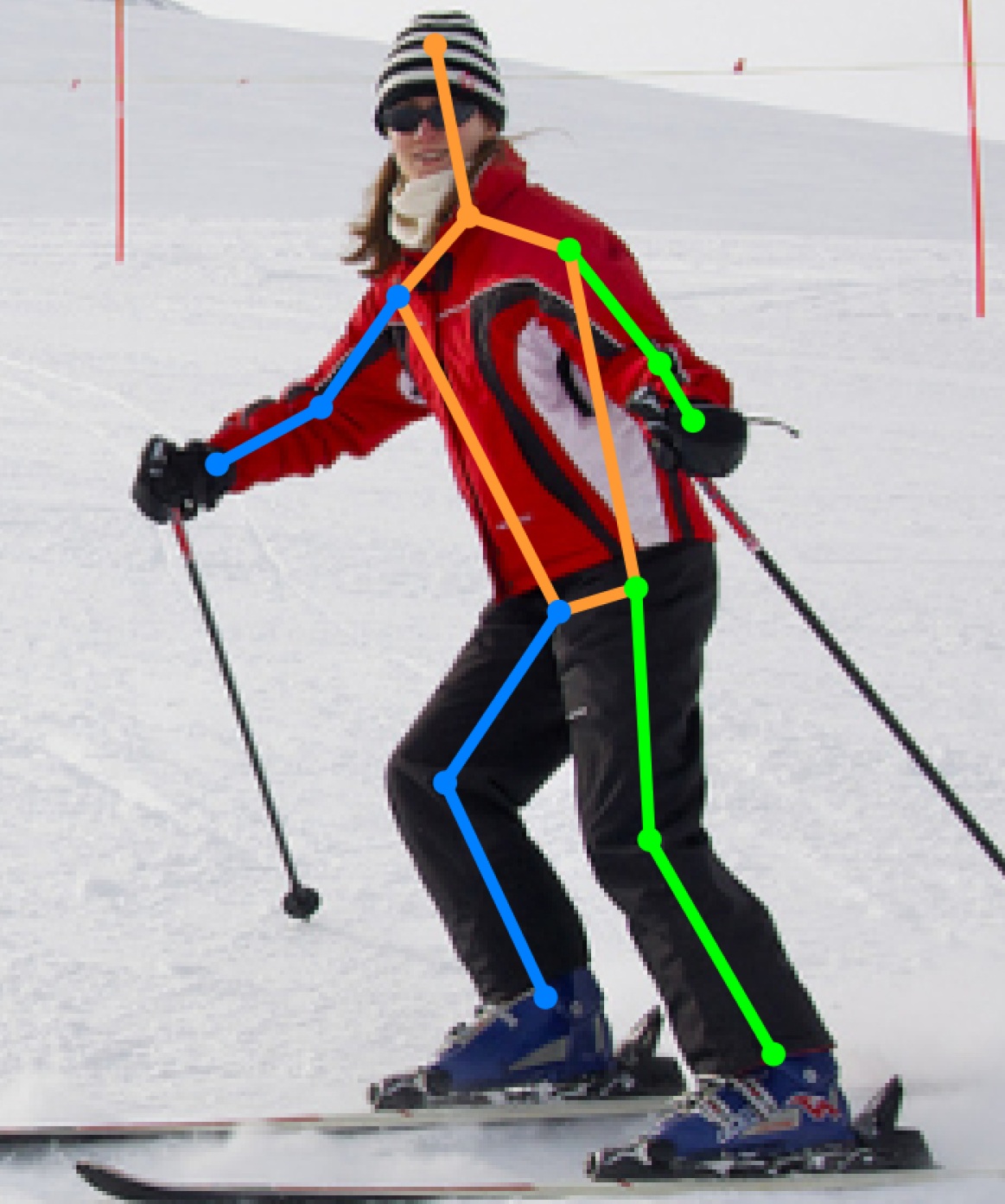} &
\includegraphics[width=0.24\linewidth]{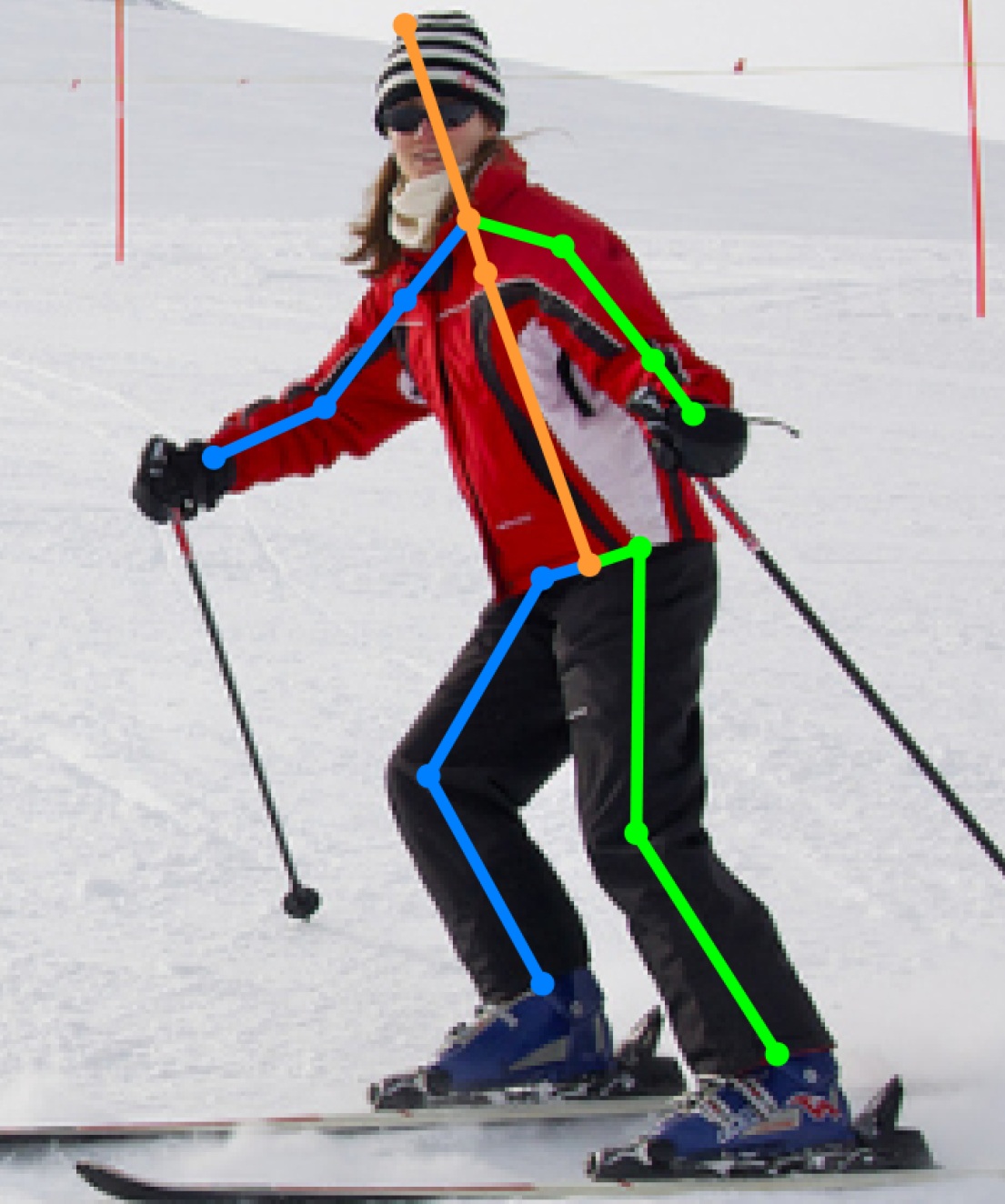} &
\includegraphics[width=0.24\linewidth]{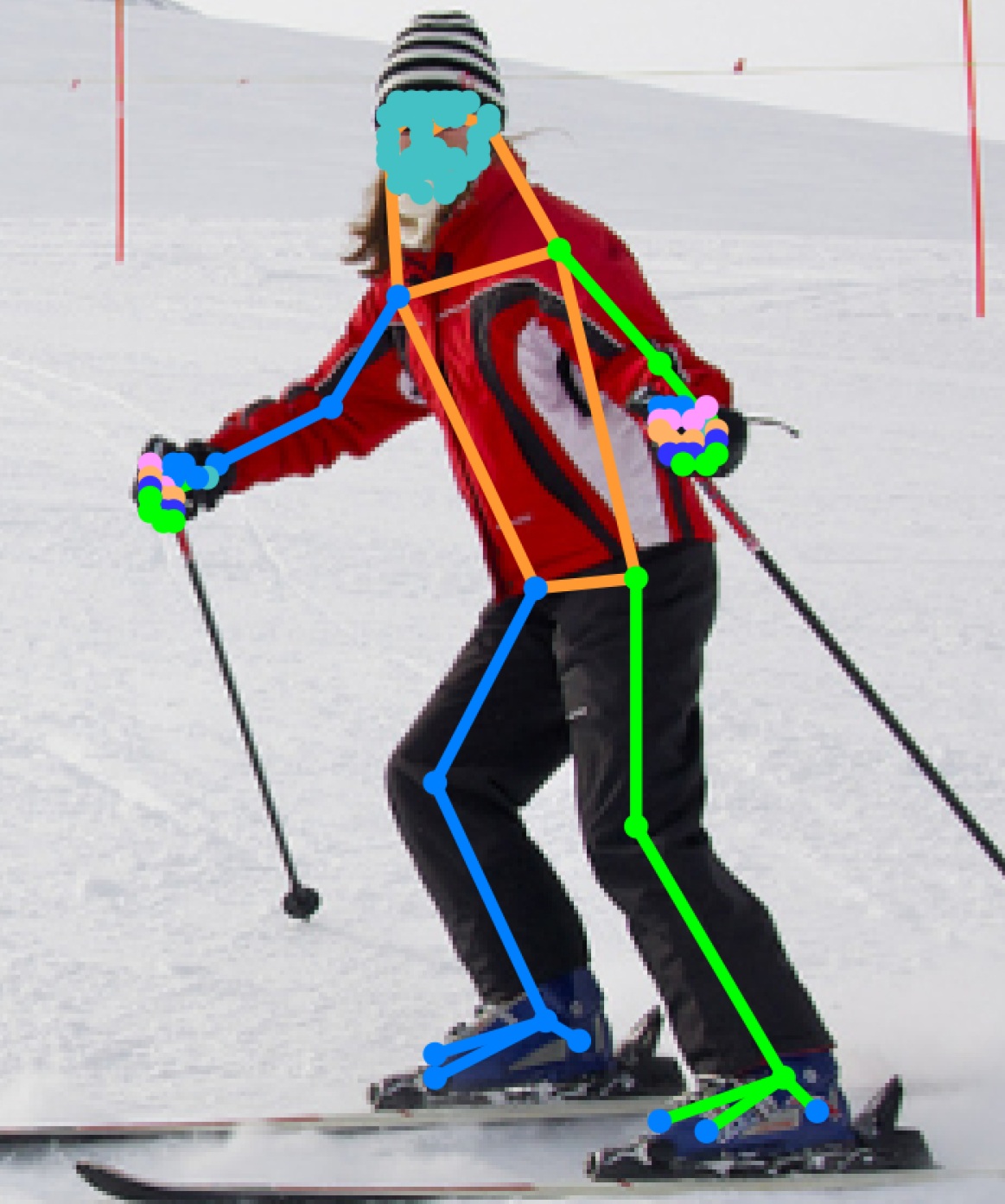} \\
COCO & AIC & MPII & WholeBody \\
\end{tabular}
\begin{tabular}{cc|c|c}
\includegraphics[width=0.24\linewidth]{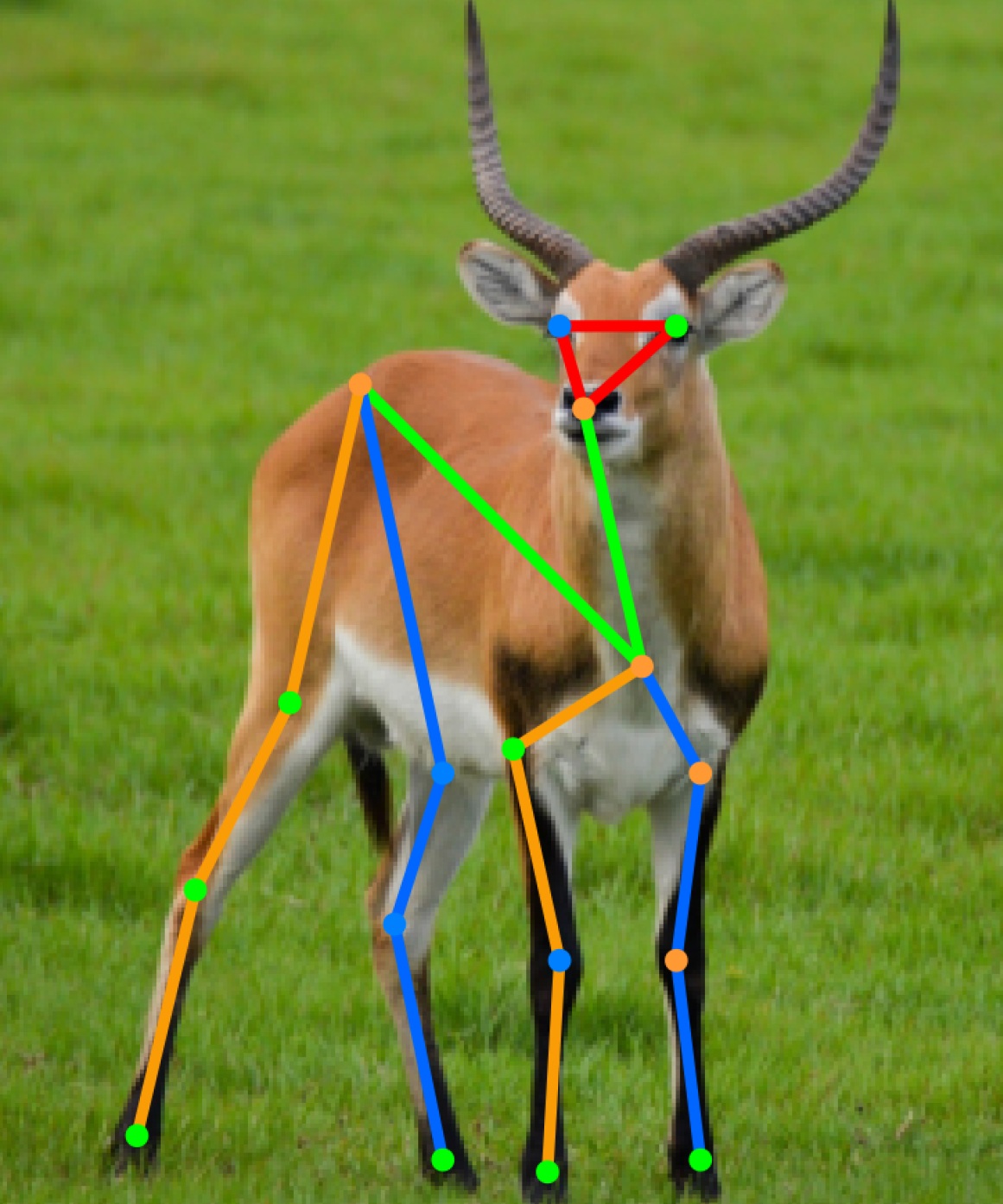} &
\includegraphics[width=0.24\linewidth]{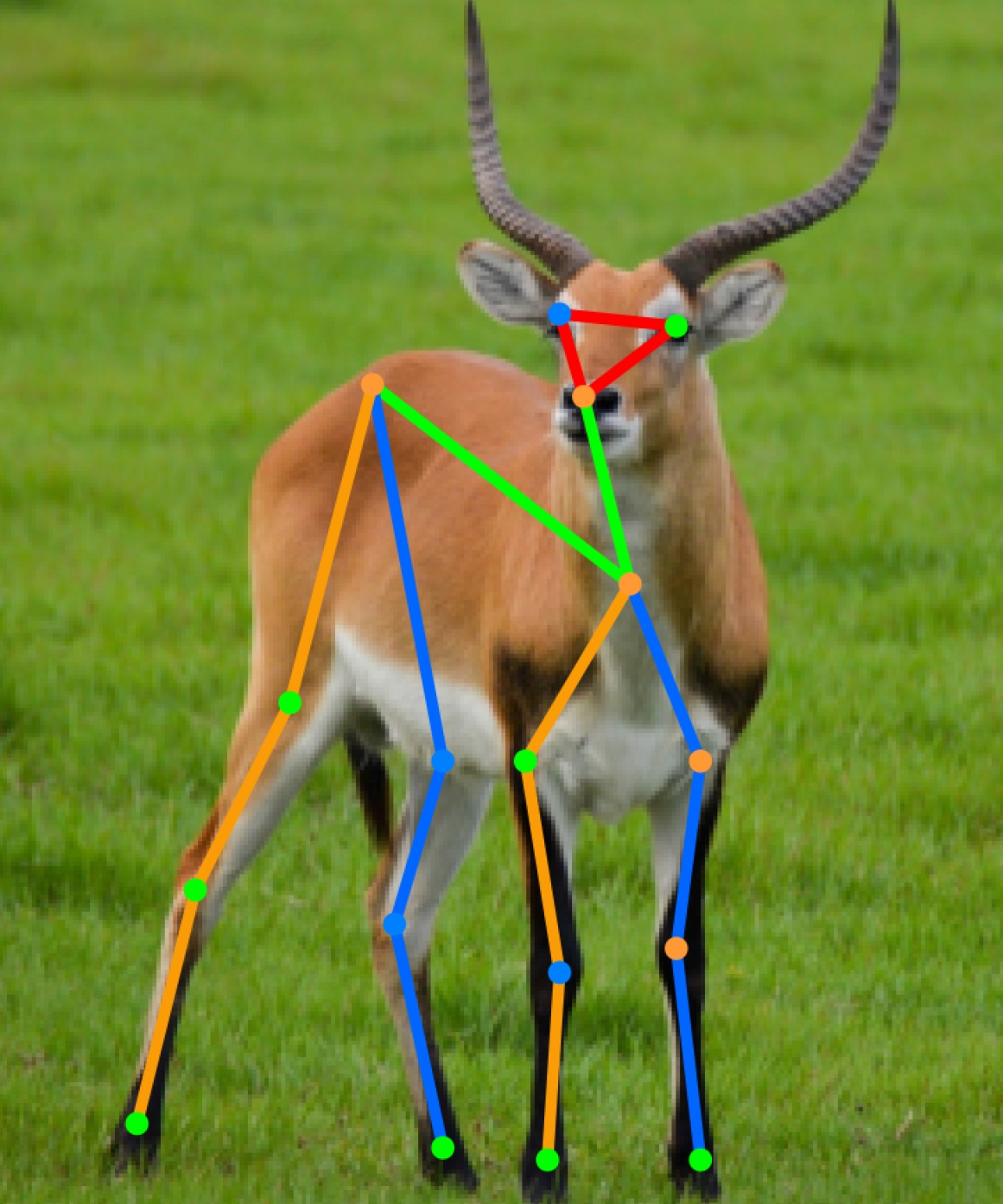} &
\includegraphics[width=0.24\linewidth]{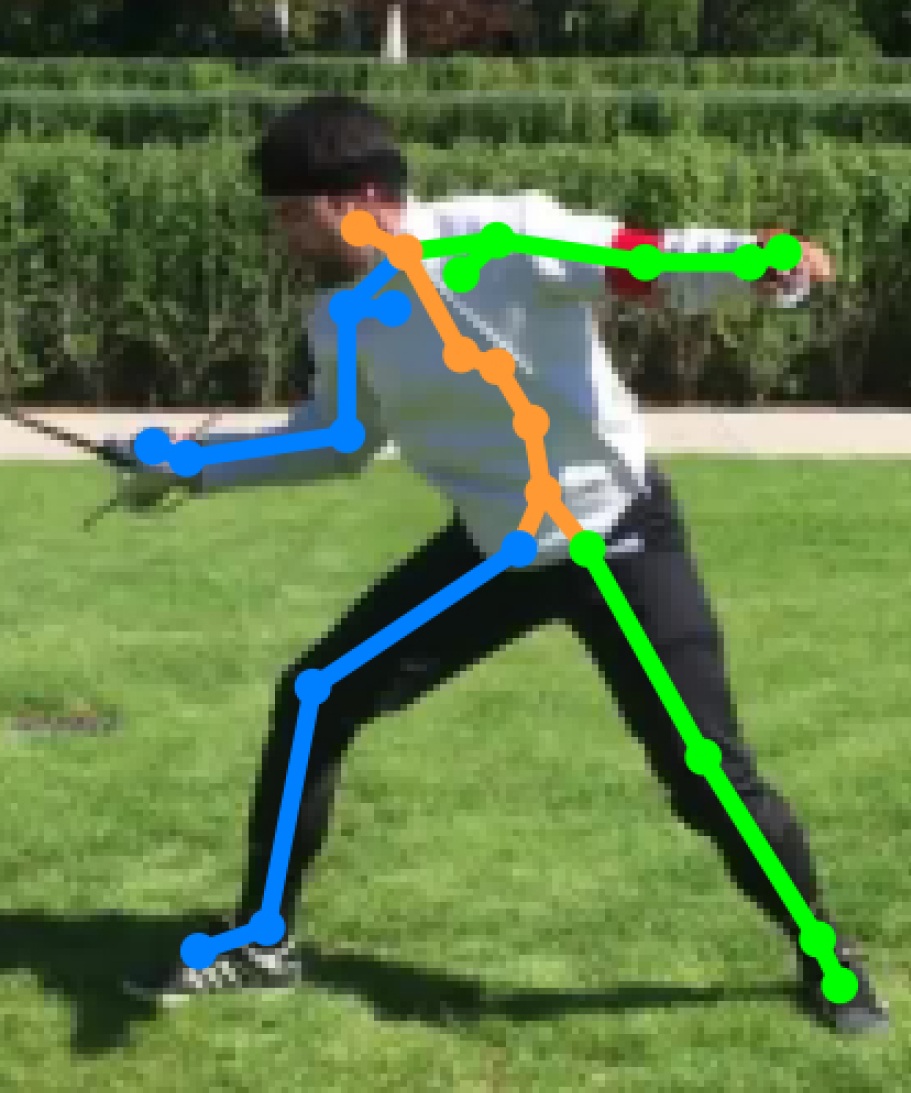} &
\includegraphics[width=0.24\linewidth]{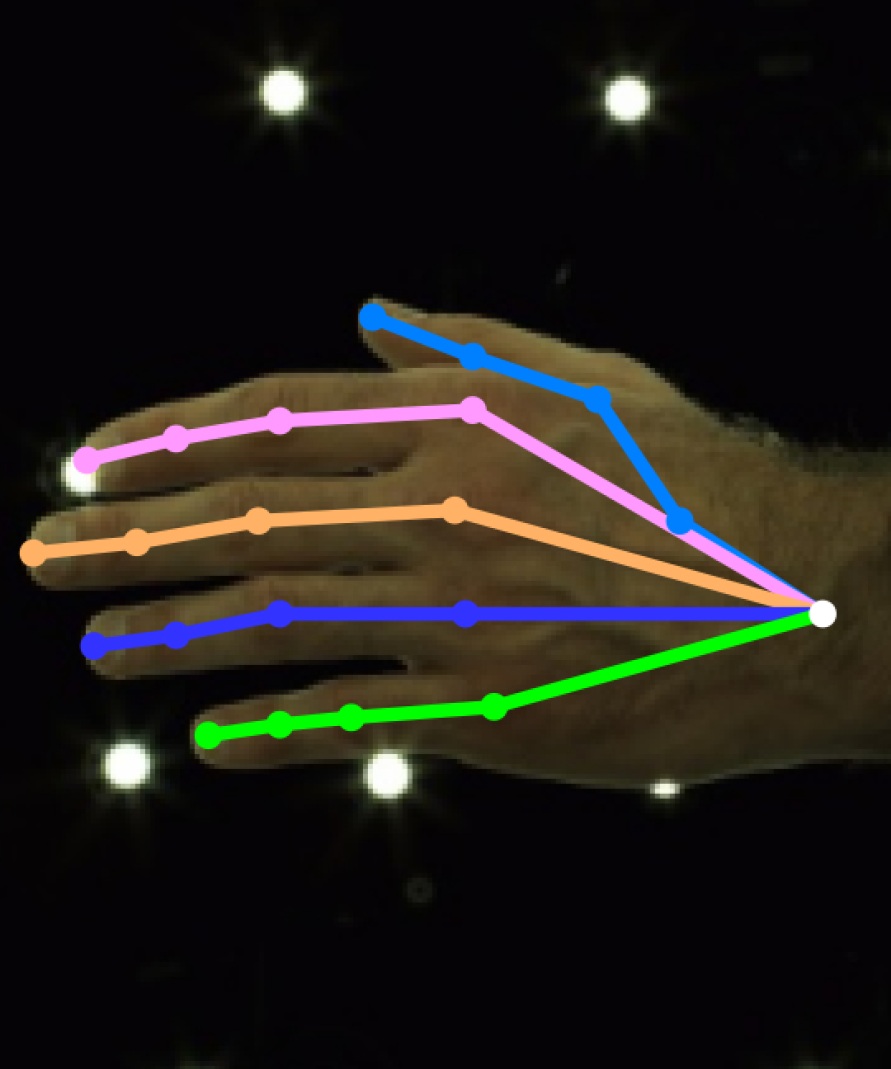} \\
AP-10K & APT-36K & 3DPW & InterHand2.6M \\
\end{tabular}
}%
\caption{PoseBH unifies diverse skeleton formats, including humans, hands, and animals. The displayed skeletons show pose estimation results from our method, with 3DPW and InterHand2.6M predictions from a transferred model.
\label{fig:feature_graphic}}
\end{figure}

However, multi-dataset training (MDT) for pose estimation presents two key challenges. First, keypoints are heterogeneous across different datasets. For example, COCO~\cite{lin2014microsoft} and MPII~\cite{andriluka14cvpr} define 17 and 16 keypoints, respectively, with 9 non-overlapping ones. Even when datasets share keypoints with identical names, their localization characteristics often differ due to domain gaps. For example, while COCO and animal datasets (\eg, AP-10K~\cite{yu2021ap} and APT-36K~\cite{yang2022apt}) use common keypoint names such as eyes, nose, and hips, significant anatomical differences exist (see \cref{fig:feature_graphic}). As a result, na\"ively merging the identically named keypoints is ineffective.

Second, pose estimation MDT suffers from a lack of adequate supervision. In MDT, multiple skeleton types from different datasets are trained for each input instance. However, only a single skeleton-type ground truth is provided per input, leaving others unsupervised. This limitation resembles a semi-supervised learning challenge, constraining performance on smaller datasets. Existing semi-supervised learning methods typically rely on input augmentation or teacher-student knowledge distillation, both of which require substantial extra computational overhead.

Existing approaches fail to fully address these challenges, often relying on a shared backbone network~\cite{ci2023unihcp} or a mixture of experts~\cite{xu2023vitposepp} to preserve downstream performance. However, these methods tend to focus on mainstream datasets like COCO or MPII, resulting in degraded performance on underrepresented datasets. We propose a method that enhances performance in discrepant domains while maintaining accuracy on standard human pose datasets. A straightforward fine-tuning approach on the target domain often disrupts performance on source datasets, as it fails to balance differences in skeleton structures and dataset distributions. Therefore, effective generalization requires jointly harmonizing skeletal representations and dataset diversity during training.

We achieve this through two primary techniques. First, we introduce prototypes to represent keypoints as embedding vectors, providing a unified representation across datasets. We regress pixel-wise keypoint embeddings from backbone features and match them to corresponding prototypes, automating keypoint mapping across datasets. 
This enables the integration of diverse keypoint types, including whole-body and animal keypoints, despite significant skeletal differences. Furthermore, our keypoint prototypes are non-learnable parameters, offering adaptability and computational efficiency for domain transfer.

Second, we generate reliable self-supervision signals using the predicted keypoints and their embeddings, a process we term \emph{cross-type self-supervision}. Since keypoint predictions can be obtained via a dot product between embeddings and prototypes, we effectively establish two distinct keypoint prediction modalities. By aligning these predictions, we filter out noisy keypoints and produce reliable labels for supervising unlabeled keypoints. This process enhances the utility of embeddings without requiring additional networks or input augmentation. As a result, we provide effective supervision for underrepresented and discrepant datasets, achieving a higher degree of generalizability.

Our main technical contributions are as follows:
\begin{itemize}
\item Keypoint prototypes for learning arbitrary keypoints from multiple datasets ensuring high transferability.
\item A cross-type self-supervision mechanism that aligns keypoint regression outputs with keypoint embeddings, enriching supervision for unlabeled keypoints.
\item Strong generalization across estimation tasks, including whole-body, animal, and hand pose estimation, as well as human shape estimation.
\end{itemize}

\section{Related works}
\label{sec:related_works}
\paragraph{Multi-dataset training} 
has been widely explored in various computer vision tasks. Relabeling and pseudo-labeling are among the most straightforward yet resource-intensive approaches. For example, MSeg~\cite{Lambert_2020_CVPR} employed relabeling for segmentation, while \cite{zhao2020object} used offline pseudo-labeling for object detection. However, both methods face scalability challenges, as adding a new dataset requires reprocessing annotations for all previously processed datasets to redefine the class set, leading to a quadratic computational cost.

Several approaches have been proposed to improve scalability. In object detection, \cite{zhou2022simple,chen2023scaledet} aim to unify object class labels by merging similar categories. However, they struggle to integrate localization features of the same object due to the absence of spatial information in class logits. UniTrack learns a unified identity embedding through self-supervised learning for multiple object-tracking tasks, including human pose tracking~\cite{wang2021different}. However, it prioritizes pose propagation over estimation and relies exclusively on the PoseTrack18~\cite{andriluka2018posetrack} skeleton, without incorporating multiple pose datasets.

Another class of methods~\cite{wang2019towards,xu2020universal,zhang2023uni3d,yang2021unik} employs network modules, such as graph convolutional networks (GCN), for cross-dataset adaptation. However, these dataset-specific modules limit domain transferability, as their input and output sizes are dictated by the label structure, making them incompatible with datasets that have different keypoint formats. In contrast, our approach achieves flexible transferability without relying on dataset-specific domain adaptation modules.

\begin{figure*}[t]
\centerline{\includegraphics[width=\textwidth]{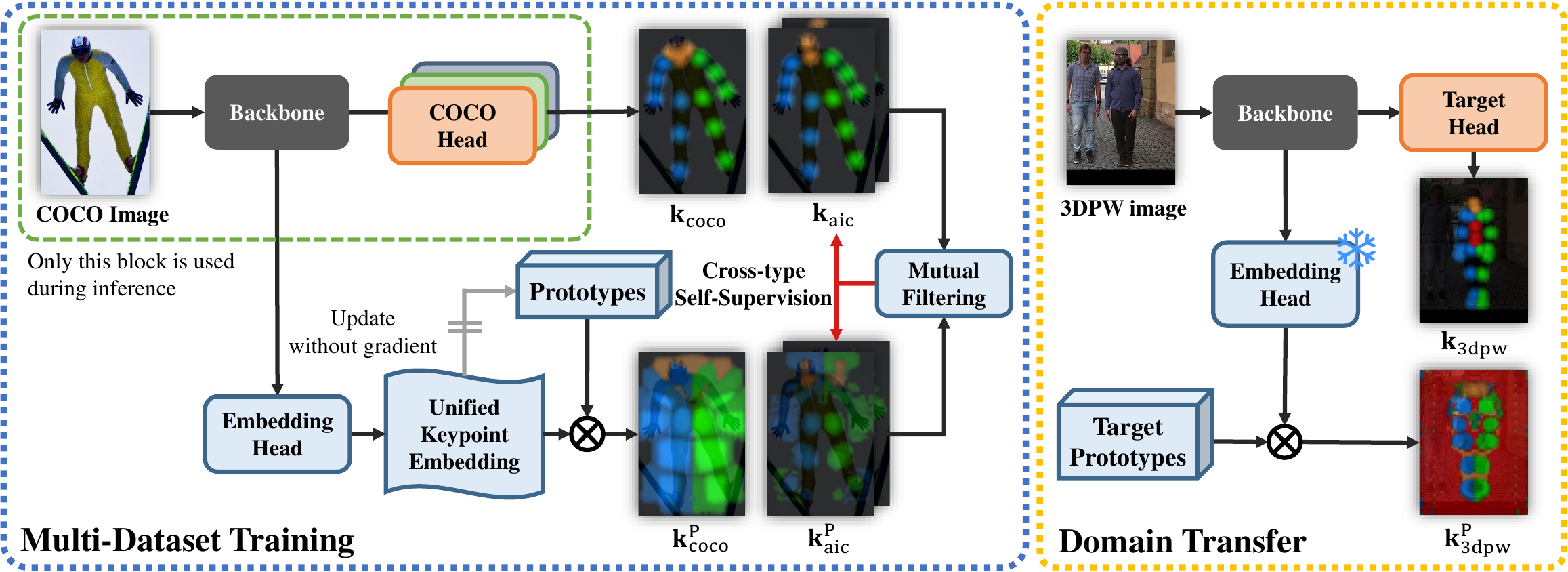}}
\caption{Overview of the PoseBH architecture. During training, the embedding head maps the backbone features into a unified keypoint embedding space. By matching these embeddings with prototypes, dataset-specific keypoint heatmaps $\mathbf{k}^{\text{P}}_{\text{coco}}$ are generated. Prototypes are updated nonparametrically from the embeddings. During inference, the embedding head and the subsequent procedures are removed.
}
\label{fig:pipeline}
\end{figure*}
\paragraph{Human pose estimation with MDT.}
In 3D human shape estimation, several methods have used pseudo-ground truths (GTs) generated by the SMPL model for MDT~\cite{ROMP, BEV, Kocabas_PARE_2021, choi2022learning}. However, these pseudo-GTs are often inaccurate due to depth ambiguity in 2D images and the limited expressiveness of the parametric 3D model~\cite{Moon_2022_CVPRW_NeuralAnnot,Moon_2023_CVPRW_3Dpseudpgts}. Similar to \cite{zhou2022simple}, S\'ar\'andi~\etal's affine-combining autoencoder (ACAE) merges 3D joint coordinates via an autoencoder~\cite{Sarandi2023dozens}. However, the autoencoder’s learnable parameters are dataset-dependent, limiting its generalizability. Furthermore, ACAE does not preserve pixel-level features, making it unsuitable when spatial information is crucial.

Recent works on category-agnostic pose estimation integrate various objects, such as chairs and vehicles, by combining multiple datasets~\cite{xu2022pose,hirschorn2023pose}. 
These methods follow a relabeling approach, creating a superset of keypoint classes and detecting only the keypoints defined in their relabeled dataset. As a result, they do not address keypoint heterogeneity or label sparsity issues, as training and evaluation occur within a single labeled dataset. In contrast, our method uses multiple skeletons to improve generalization across diverse downstream domains.

In 2D human pose estimation, UniHCP performs multiple human perception tasks using task-specific queries without incorporating task-specific layers inside the backbone~\cite{ci2023unihcp}. This design limits its ability to generalize to distant domains such as animals and whole-body pose estimation. ViTPose++~\cite{xu2023vitposepp} and Sapiens~\cite{khirodkar2024sapiens} employ multi-head architectures for MDT but do not address skeleton heterogeneity or the resulting label sparsity.

\paragraph{Prototype learning} can be interpreted as proxy-based metric learning~\cite{roth2022non}, where the centers of a fixed set of clusters serve as prototypes. This approach has been studied in various domains, including semantic segmentation~\cite{du2022weakly,zhou2022rethinking,das2023weakly}, person re-identification~\cite{rao2023transg}, meta-learning~\cite{snell2017prototypical,yoon2019tapnet,Zhang_2021_CVPR,liu2023learning}, and visual interpretability~\cite{chen2019looks,donnelly2022deformable,Wang_2023_ICCV}.

ProtoSeg~\cite{zhou2022rethinking} introduces a nonparametric approach 
to semantic segmentation by assigning multiple prototypes per class. ProMotion~\cite{lu2024promotion} employs prototypes to unify scene depth estimation and optical flow tasks. While our method draws inspiration from ProtoSeg and aligns with ProMotion in exploiting the benefits of prototypes for MDT, it differs in three key aspects. First, our approach applies keypoint-level supervision, which is inherently sparser than segmentation, depth, or optical flow. Second, we specifically address skeleton heterogeneity and cross-dataset relationships, aspects not considered by ProtoSeg and ProMotion. Third, while ProtoSeg and ProMotion apply fully supervised learning, our method tackles the challenge of unlabeled data, a problem unique to MDT pose estimation.

\section{Method}
We focus on multi-dataset training (MDT) for 2D human pose estimation. Given $D$ datasets, our model predicts a keypoint heatmap for a person, represented as an array $\mathbf{k}_{d}\in\mathbb{R}^{J_d\times H\times W}$, where $d$ is the dataset index, $J_d$ is the number of keypoints in the $d$-th dataset, and $H$ and $W$ denote the height and width of the heatmap, respectively.

\subsection{Multi-head baseline}
The multi-head baseline model consists of a backbone network that extracts an output backbone feature map from an input RGB image. 
The backbone network is shared across all datasets, while individual keypoint heads independently regress dataset-wise keypoint heatmaps $\{\mathbf{k}_{d}\}_{d=1}^D$, each trained with its corresponding labels. The keypoint heads share the same architecture, except for the final layer, where the output channel size is the respective number of keypoints. This generic baseline does not share the keypoint output space, limiting the MDT effect up to the backbone.

\subsection{Keypoint prototype learning}
\paragraph{Overview.} Our goal is to learn a unified keypoint space that standardizes keypoint output formats across datasets, scales with dataset size, and retains dataset-specific localization accuracy. To achieve this, we reformulate keypoint regression as a prototype-based distance metric learning problem (see~\cref{fig:pipeline}). Our shared keypoint embedding module generates a normalized keypoint embedding map $\mathbf{e}\in\mathbb{R}^{F\times H\times W}$ from an input feature map. The module consists of two deconvolution layers, followed by a residual block and two convolution layers. Additional details are provided in the supplementary document.

Using embeddings and prototypes, we perform keypoint classification by matching embeddings with prototypes $\mathbf{P}\in\mathbb{R}^{J\times M\times F}$, where $F$ is the embedding dimension, $J$ is the total number of keypoints across all datasets, $M$ is the number of in-class prototypes. For each heatmap pixel, we compute the cosine similarity between the prototypes and the embedding vector to obtain keypoint predictions:
\begin{align}
\mathbf{k}^\text{P}_{j,y,x} = \max_{m} \frac{\mathbf{P}_{(j,m,:)}\cdot\mathbf{e}_{(:,y,x)}}{\left\|\mathbf{P}_{(j,m,:)}\right\|\left\|\mathbf{e}_{(:,y,x)}\right\|},
\label{eq:proto_match}
\end{align}
where $x,y$ denote pixel indices in the heatmap, 
and $\mathbf{P}_{(j,m,:)}$ is a row vector of size $F$, extracted $\mathbf{P} \in \mathbb{R}^{J \times M \times F}$ along its first two dimensions. For each keypoint class $j$, $\mathbf{k}^\text{P}_{(j,:,:)}\in \mathbb{R}^{H\times W}$ stores the matching score of the most probable prototype among $M$ prototypes. We refer to this as the \emph{prototype keypoint heatmap} $\mathbf{k}^\text{P}$ to distinguish it from the multi-head outputs $\mathbf{k}$. 

For the remainder of this paper, we use the following notation: $\mathbf{A}_{(:,:,a)}$ denotes the 2D slice of a 3D array $\mathbf{A}$ indexed at $a$ along the third dimension, while $\mathbf{A}_{(:,b,:)}$ represents the 2D slice at index $b$ along the second dimension. Similarly, $\mathbf{A}_{(d:e,:,:)}$ represents the 3D sub-array spanning indices $d$ to $e$ along the first dimension. For a 2D array $\mathbf{B}$, $\mathbf{B}_{(:,c)}$ indicates the $c$-th column vector, whereas $\mathbf{B}_{(d,c)}$ refers to the element at position $(d,c)$.

\paragraph{Prototype learning.} 
We adopt a nonparametric learning approach~\cite{zhou2022rethinking} to train $\mathbf{P}$. First, we obtain logits $\{\mathbf{l}_j\}_{j=1}^J \subset \mathbb{R}^{M \times N}$ and targets $\{\mathbf{t}_j\in \mathbb{R}^{N_j}\}_{j=1}^J$, where $N_j$ is the number of foreground samples (\ie, pixels in the ground-truth (GT) heatmap with nonzero values) for the $j$-th joint, and $N$ is the total number of foreground samples across all joints. We define logits as sampled embedding vectors and targets as keypoint class labels, with each keypoint class selecting an in-class prototype from $M$ candidates.

Since no GT selection is explicitly provided, we assign an in-class prototype for each logit via online clustering using Sinkhorn-Knopp iteration with an equipartition constraint~\cite{SP67}, and compute $\mathbf{t}_j$ as follows:
\begin{align}
\label{eq:logit}
\mathbf{l}_j &= \mathbf{P}_{(j,:,:)} \overline{\mathbf{e}}, \\
\mathbf{t}_j &= \argmax_{m} \diag (\mathbf{u}) \exp \left( \frac{\mathbf{l}_j}{\kappa} \right) \diag (\mathbf{v}),
\label{eq:target}
\end{align}
where $\overline{\mathbf{e}}$ is a 2D array of size $F \times N_j$, obtained by first flattening $\mathbf{e}\in\mathbb{R}^{F\times H\times W}$ along the spatial dimensions into a 2D array of size $F\times (HW)$ and then, selecting $N_j$ foreground objects. $\diag(\mathbf{u})$ is a diagonal matrix derived from the vector $\mathbf{u}$, and $\exp(\cdot)$ is applied element-wise.

Each prototype is then updated with momentum $\lambda$:
\begin{equation}
\mathbf{P}_{j,m}^{\text{new}} = \lambda \mathbf{P}_{j,m}^{\text{old}} + (1-\lambda) \frac{1}{N_{j}}\sum_{n=1}^{N_{j}} w_{mn} \overline{\mathbf{e}}_{(:,n)},
\label{eq:proto_update}
\end{equation}
where $w_{mn}$ is the clustering assignment weight, calculated similarly to~\cref{eq:target}, but using $\max$ instead of $\argmax$.

To optimize the learned embedding with respect to the prototype $\mathbf{P}$, we employ pixel-prototype contrastive learning, using the pixel-prototype contrastive loss (PPC) and pixel-prototype distance (PPD) from~\cite{zhou2022rethinking}:
\begin{align}
\mathcal{L}_{\text{PPC}}(\mathbf{l}_j,\mathbf{t}_j) &= \frac{c_n}{N_j}\sum_{n=1}^{N_{j}}\text{CE}\left(\mathbf{l}_{j,n},\mathbf{t}_{j,n}\right),
\label{eq:ppc}\\
\mathcal{L}_{\text{PPD}}(\mathbf{l}_j,\mathbf{t}_j) &= \frac{c_n}{N_j}\sum_{n=1}^{N_{j}}(1-\mathbf{l}_{j,n}^{m+})^2,
\label{eq:ppd}
\end{align}
where $c_n$ is the confidence value from the GT heatmap, $\text{CE}$ denotes the cross-entropy loss, and $m^{+}$ indexes the first dimension of $\mathbf{l}_d$. For each dataset $d$, we collect logits $\mathbf{l}_{j,n} \in \mathbb{R}^{M \times N_d}$ and targets $\mathbf{t}_d \in \mathbb{R}^{N_d}$, where $N_d$ is the number of samples in dataset $d$. The total prototype loss is then defined as $\mathcal{L}_{\text{Proto}} = \mathcal{L}_{\text{PPC}} + \mathcal{L}_{\text{PPD}}$.

\begin{figure}[t]
\centerline{\includegraphics[width=\columnwidth]{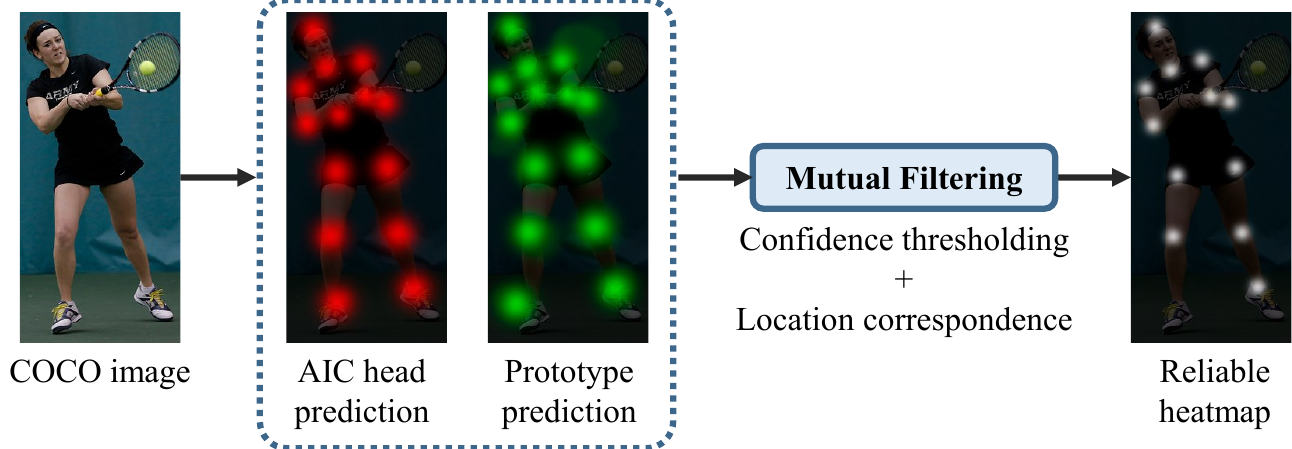}}
\caption{An illustrative example of cross-type self-supervision. Given a COCO image, we jointly refine the AIC head and AIC prototype predictions to produce reliable AIC heatmaps.
}
\label{fig:css_overview}
\end{figure}

Unlike standard contrastive learning, $\mathcal{L}_{\text{Proto}}$ operates within a single dataset and lacks the push term for differing dataset skeletons. To introduce cross-dataset negatives, we apply K-means clustering to all prototypes, forming $K$ (=96) clusters. This clustering is performed only once during training, while online clustering for in-class prototypes is conducted at each iteration. These clusters serve as negative samples, where the embeddings closest to each centroid are used as logits, with their dataset-specific labels as targets. Finally, we obtain $\mathbf{l}_{c}$ and $\mathbf{t}_{c}$ for each cluster and apply \cref{eq:ppc,eq:ppd} to enable cross-dataset contrastive learning.

\subsection{Cross-type self-supervision}
While our keypoint prototypes enable unified MDT, label sparsity remains a challenge, as each image is annotated for only one dataset, limiting supervision. To address this semi-supervised learning problem, we introduce a self-supervision strategy that bridges keypoint heatmaps and embeddings. By collaboratively filtering out noisy predictions from both the keypoint and embedding heads, we generate reliable keypoint heatmaps for training (see~\cref{fig:css_overview}).

To eliminate uncertain keypoint predictions, we apply two filtering conditions. First, for each keypoint class, the confidence scores of both the keypoint and embedding heads must exceed a threshold ($c_{thr}=0.25$). Second, the root mean square distance between their predictions must be below a threshold ($d_{thr}=2.1$). 

The filtered predictions are then combined using a weighted average:
\begin{align}
\widehat{\mathbf{y}}_{i} = s_{i}\widehat{\mathbf{y}}_{i}^{\text{kpt}} + (1-s_{i})\widehat{\mathbf{y}}_{i}^{\text{emb}},
\label{eq:ssp_pred}
\end{align}
where $s_{i} = c^{\text{kpt}}_{i}/(c^{\text{kpt}}_{i}+c^{\text{emb}}_{i})$, $i$ is the keypoint index, and $\widehat{\mathbf{y}}$ and $c$ represent the keypoint prediction and confidence score from each head, respectively.

\begin{table}[!t]
\centering
{
\footnotesize
\begin{tabular}{l|c|ccc|c}
\toprule
Method & GFLOPS & COCO & AIC & MPII & \textit{Avg.} \\ \midrule
UniHCP & 18.9 & 76.8 & \textbf{32.6} & 90.9 & 66.8 \\
ViTPose++ & 18.5 & 77.0 & 31.6 & 93.1 & 67.2 \\ \midrule
\rowcolor[gray]{.8}
Ours & 18.5 & \textbf{77.3} & 32.1 & \textbf{93.2} & \textbf{67.5} \\
\bottomrule
\end{tabular}
}
\caption{Comparison of different multi-dataset training methods on general human pose benchmarks, with computational complexity measured in GFLOPS during the evaluation phase.}
\label{table:sota_mdt}
\end{table}
\begin{table}[t]
\centering
\begin{tabular}{l|ccc|c}
\toprule
Method & AP-10K & APT-36K & COCO-W & \textit{Avg.} \\ \midrule
UniHCP & 56.5 & 62.0 & 20.1 & 46.2 \\
ViTPose++ & 74.1 & 76.0 & 57.1 & 69.1 \\ \midrule
\rowcolor[gray]{.8}
Ours & \textbf{75.0} & \textbf{87.2} & \textbf{57.9} & \textbf{73.4} \\
\bottomrule
\end{tabular}
\caption{Comparison of multi-dataset training methods on whole-body and animal domains. COCO-W refers to COCO-WholeBody. The reported metrics are AP(\%).}
\label{table:animal_transfer}
\end{table}

From $\widehat{\mathbf{y}}$, we generate a \emph{reliable heatmap} $\mathbf{k}^{\text{CSS}}$ following the standard GT heatmap generation process. The loss for unlabeled samples is then computed as: 
\begin{align}
\mathcal{L}_{\text{CSS}} &= \sum_{d=1}^{D} \zeta \big[ \mathcal{L}_{\text{hm}}(\mathbf{k}_{d[\mathbf{u}]},\mathbf{k}^{\text{CSS}}_{d[\mathbf{u}]}) \nonumber\\
&+  \mathcal{L}_{\text{Proto}}(\mathbf{e}_{[\mathbf{u}]},\mathbf{k}_{d[\mathbf{u}]},\mathbf{k}^{\text{CSS}}_{d[\mathbf{u}]}) \big],
\label{eq:CSS}
\end{align}
where $\mathbf{u}$ denotes the indices of the unlabeled samples. 
This cross-type self-supervision (CSS) loss enables self-distillation between the dataset-wise keypoint head and the prototype, mitigating supervision shortage without requiring a teacher model or input duplication.

The final training loss is defined as:
\begin{align}
\mathcal{L}_{\text{MDT}}&=\mathcal{L}_{\text{KPL}}+\mathcal{L}_{\text{CSS}}, \text{ where }\\ 
\mathcal{L}_{\text{KPL}}&= \sum_{d=1}^{D} \left[ \mathcal{L}_{\text{hm}}(\mathbf{k}_{d},\mathbf{k}^{\text{gt}}_{d}) + \mathcal{L}_{\text{Proto}}(\mathbf{e},\mathbf{k}_{d},\mathbf{k}^{\text{gt}}_{d})\right],
\end{align}
and $\mathcal{L}_{\text{hm}}$ is the standard JointMSE loss~\cite{sun2019deep}.

\begin{table*}[!t]
\centering
{
\begin{tabular}{l|l|r|cccc|cc}
\toprule
\multirow{2}{*}{Method} & \multirow{2}{*}{Backbone} & \multirow{2}{*}{Datasets} & \multicolumn{4}{c|}{COCO} & \multicolumn{2}{c}{MPII} \\
 & & & (AP$\uparrow$) & (AP$^{50}\uparrow$) & (AP$^{75}\uparrow$) & (AR$\uparrow$) & (PCKh$\uparrow$) & (PCKh@0.1$\uparrow$) \\
\midrule 
HRNet & HRNet-W48 & 1 & 75.1 & 90.6 & 82.2 & 80.4 & \hspace{1.3mm}90.3$^{\dagger}$ & \hspace{1.3mm}33.1$^{\dagger}$ \\
HRFormer & HRFormer-B & 1 & 75.6 & 90.8 & 82.8 & 80.8 & - & - \\
SimCC & HRNet-W48 & 1 & 76.1 & 90.6 & 82.9 & 81.2 & 90.0 & 36.8 \\
PCT$^{\dagger}$ & Swin-B & 1 & 77.7 & 91.2 & 84.7 & 82.1 & 92.5 & - \\
PCT$^{\dagger}$ & Swin-H & 1 & 79.3 & 91.5 & 85.9 & - & - & - \\ \midrule
UniHCP & ViT-B & 33 & 76.1 & - & - & - & \hspace{2.6mm}93.2$^{\dagger\dagger}$ & -  \\
ViTPose++-B & ViT-B & 6 & 77.0 & 73.4 & 84.0 & 82.6 & 92.8 & 39.1  \\ 
ViTPose++-H & ViT-H & 6 & 79.4 & \textbf{91.9} & 85.7 & \textbf{84.8} & \textbf{94.2} & 41.6 \\
\midrule
\rowcolor[gray]{.8}
Ours & ViT-B & 6 & 77.3 & 90.8 & 84.2 & 82.4 & 93.2 & 39.3 \\
\rowcolor[gray]{.8}
Ours & ViT-H & 6 & \textbf{79.5} & \textbf{91.9} & \textbf{85.8} & 84.5 & \textbf{94.2} & \textbf{42.0} \\
\bottomrule
\end{tabular}
}%
\caption{Comparison with state-of-the-art methods on COCO and MPII. Methods with $^{\dagger}$ uses 256$\times$256 input size. $^{\dagger\dagger}$ indicates additional fine-tuning on each dataset.}
\label{table:sota_co_mp}
\end{table*}

\section{Experiments}
\subsection{Experimental setup}
We follow the experimental setup of ViTPose++~\cite{xu2023vitposepp}, using six human pose datasets for training.
(1) COCO consists of 57,000 training and 5,000 validation images, annotated with 17 keypoints~\cite{lin2014microsoft}. 
(2) AIChallenger (AIC) contains 210,000 training and 30,000 validation images, annotated with 14 keypoints~\cite{wu2017ai}. 
(3) MPII includes approximately 15,000 training, 2,729 validation, and 5,700 test images, with 16 keypoints~\cite{andriluka14cvpr}. 
(4) AP-10K is an animal pose dataset with approximately 7,000 training, 1,000 validation, and 2,000 test images~\cite{yu2021ap}. 
(5) APT-36K, a video-based animal pose dataset, consists of about 25,000 training, 3,600 validation, and 7,000 test images~\cite{yang2022apt}. Both AP-10K and APT-36K share the same skeletal format with 17 keypoints, largely following COCO. (6) COCO-WholeBody extends COCO with whole-body annotations covering 133 keypoints~\cite{wu2017ai}.
We follow standard evaluation protocols, using the average precision (AP) metric based on object keypoint similarity (OKS) for COCO, AIC, COCO-WholeBody, AP-10K, and APT-36K. For MPII, the percentage of correct keypoints (PCK) metric is used~\cite{andriluka14cvpr}.

For the domain transfer experiments, we use InterHand2.6M~\cite{Moon_2020_ECCV_InterHand} and 3DPW~\cite{vonMarcard2018}. InterHand2.6M is a two-hand pose estimation dataset containing approximately 2.6M frames, annotated with 21 keypoints. We adopt the ViTPose++ configuration for frame sampling and dataset splits (training, validation, and test). For evaluation, we use PCK, AUC, and EPE, consistent with ViTPose++. 3DPW is an in-the-wild 3D human shape estimation dataset with 18,000 training and 26,000 test frames, annotated with 24 keypoints. We use COCO-style AP and AR as evaluation metrics. In both InterHand2.6M and 3DPW, the projected 2D keypoints are used as ground-truths.

Further details on the experimental setups, the impact of varying hyperparameters, and failure case analysis can be found in the supplemental document.

\subsection{Implementation details}
Our multi-head baseline model is ViTPose++~\cite{xu2023vitposepp}. We use an input image size of 256$\times$192 and apply a flip test in all cases. The model is trained for 100 epochs using the AdamW optimizer with a weight decay factor of 0.1 and a step learning rate scheduler with an initial learning rate of 0.001. The learning rate is reduced by a factor of 0.1 at the 50th and 90th epochs. We use four NVIDIA A100 GPUs for training with the ViT-H backbone and four NVIDIA A6000 GPUs for training with the ViT-B backbone.

We employ a scheduling strategy for both the losses and the learning rate. Initially, we freeze the multi-head model, updating only the embedding module and prototypes for 50 epochs. In the subsequent 40 epochs, we train the multi-heads and the embedding module while keeping the prototypes and the backbone frozen. Finally, during the last 10 epochs, we train the entire network except the prototypes. At this stage, we introduce cross-type self-supervision loss. This progressive optimization approach accelerates training compared to training the entire network from the outset.

\subsection{Comparison with existing MDT methods}
In~\cref{table:sota_mdt}, we evaluate the competitiveness of our method against existing MDT approaches. For a fair comparison, all methods use the ViT-Base backbone and are trained under our dataset configuration. While UniHCP performs well on AIC, it achieves the lowest scores on COCO and MPII, resulting in an average of 66.8. We hypothesize that UniHCP's parameter-sharing strategy causes well-represented pose datasets to dominate, hindering its generalization across different skeleton structures. ViTPose++ serves as both a multi-head model and the baseline for our method. Our algorithm surpasses ViTPose++ by 0.3 in average score and outperforms it on individual datasets. Given that these benchmark scores are highly saturated and our method maintains the same computation complexity as ViTPose++ during inference, this improvement is significant.

To further assess generalization, we test our method on animal and whole-body pose estimation datasets. \Cref{table:animal_transfer} shows that our method consistently improves performance, despite significant domain discrepancies. UniHCP lags behind with an average score of 27.2, likely due to its parameter sharing strategy, which does not address skeleton heterogeneity. Our method also surpasses ViTPose++ by an average score of 4.3, demonstrating superior generalization beyond human skeletons. Specifically, it achieves gains of 0.9 AP on AP-10K, 11.2 AP on APT-36K, and 0.8 AP on COCO-WholeBody. Unlike existing methods that compromise either mainstream performance or adaptation to distant domains, our approach enhances both synergistically.

\begin{figure*}[t]
\centering
\begin{tabular}{ccccc|cc}
 & COCO & AIC & MPII & COCO-W & AP-10K & APT-36K \\
ViTPose++ & \raisebox{-0.5\height}{\includegraphics[height=0.36\columnwidth]{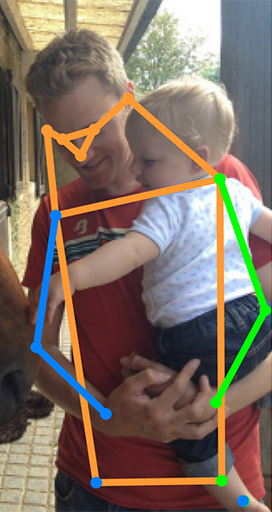}} & \raisebox{-0.5\height}{\includegraphics[height=0.36\columnwidth]{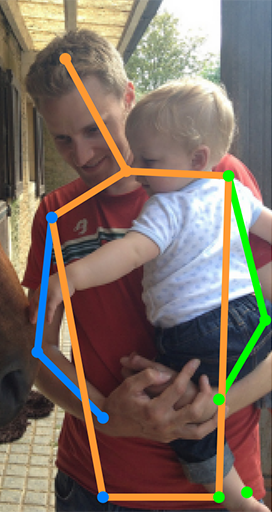}} & \raisebox{-0.5\height}{\includegraphics[height=0.36\columnwidth]{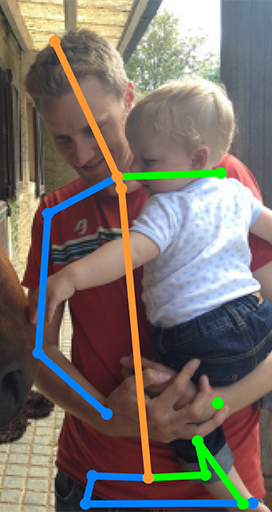}} & \raisebox{-0.5\height}{\includegraphics[height=0.36\columnwidth]{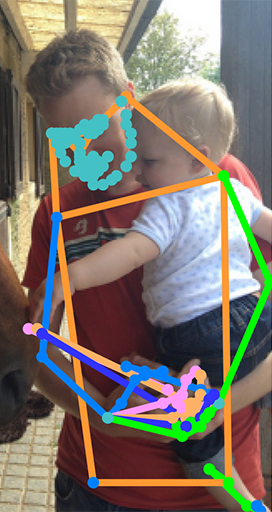}} & \raisebox{-0.5\height}{\includegraphics[height=0.36\columnwidth]{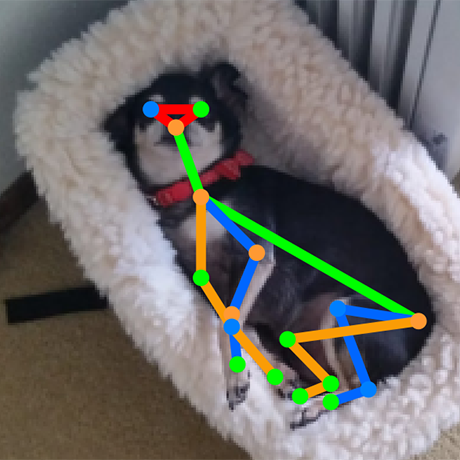}} & \raisebox{-0.5\height}{\includegraphics[height=0.36\columnwidth]{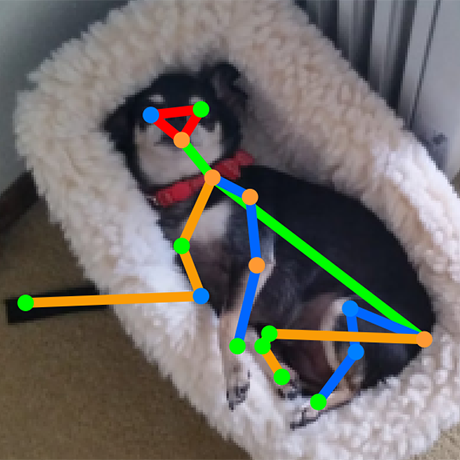}} \\
Ours & \raisebox{-0.5\height}{\includegraphics[height=0.36\columnwidth]{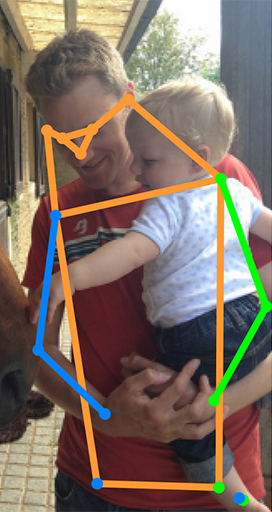}} & \raisebox{-0.5\height}{\includegraphics[height=0.36\columnwidth]{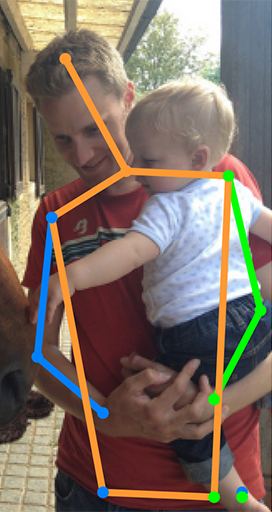}} & \raisebox{-0.5\height}{\includegraphics[height=0.36\columnwidth]{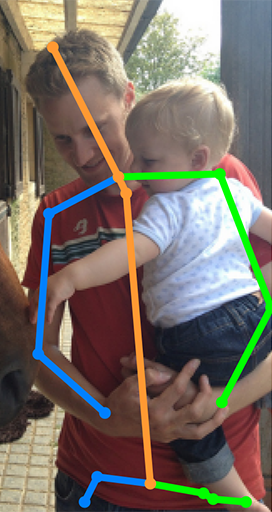}} & \raisebox{-0.5\height}{\includegraphics[height=0.36\columnwidth]{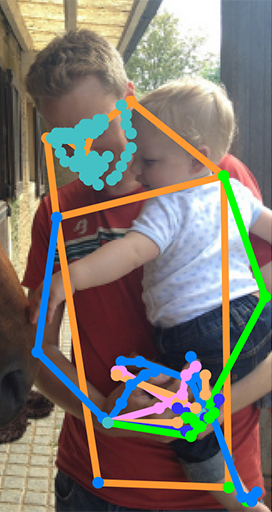}} & \raisebox{-0.5\height}{\includegraphics[height=0.36\columnwidth]{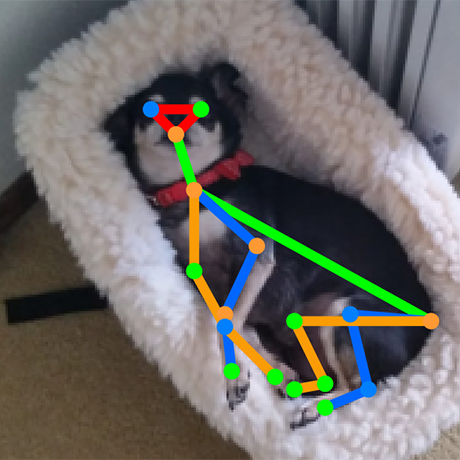}} & \raisebox{-0.5\height}{\includegraphics[height=0.36\columnwidth]{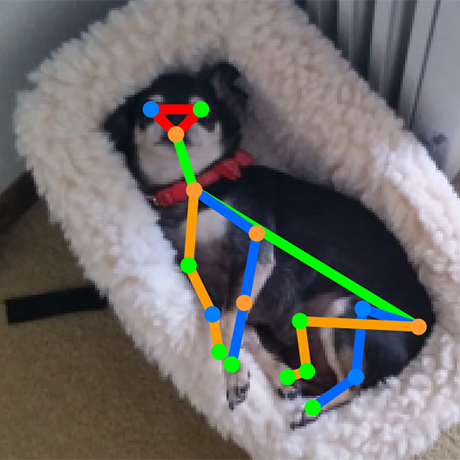}} \\
\end{tabular}
\vspace{-0.18cm}
\caption{Comparative pose estimation results on human (left) and animal (dog; right).}
\label{fig:kpt_pred}
\vspace{-0.1cm}
\end{figure*}

\subsection{Comparison to state-of-the-art methods}
In~\cref{table:sota_co_mp}, we compare our method with state-of-the-art (SOTA) approaches on the COCO and MPII benchmarks, using scores reported in the respective papers. We also evaluate our method against SOTA approaches designed for single datasets, including HRNet~\cite{sun2019deep}, HRFormer~\cite{YuanFHLZCW21}, SimCC~\cite{li2022simcc}, and PCT~\cite{Geng23PCT}. On the COCO validation set, our method ranks second with the ViT-B backbone and achieves SOTA performance with the ViT-H backbone, reaching 79.5 AP. On the MPII validation set, our ViT-B model performs comparably to the fine-tuned UniHCP and surpasses the SOTA on the PCKh@0.1 metric by 0.4 PCKh. These results demonstrate that our method effectively balances adaptability to diverse domains while maintaining competitive performance on standard benchmarks, aligning with our primary objective of improving generalization beyond mainstream tasks.

\begin{table}[]
\begin{tabular}{l|ccc|cc}
\toprule
\multirow{2}{*}{Method} & \multicolumn{3}{c|}{InterHand2.6M} & \multicolumn{2}{c}{3DPW} \\
        & PCK$\uparrow$ & AUC$\uparrow$ & EPE$\downarrow$ & AP$\uparrow$ & AR$\uparrow$ \\
\midrule
UniHCP        & 98.5 & 87.0 & 3.72 & 57.9 & 61.8 \\
ViTPose++     &   98.3        &   86.2        &  4.02         &  81.7         & 85.2       \\
\midrule
\rowcolor[gray]{.8}
Ours          & \textbf{98.6} & \textbf{87.1} & \textbf{3.70} & \textbf{83.6} & \textbf{87.1} \\
\bottomrule
\end{tabular}
\caption{Transfer results on hand (Interhand2.6M) and human shape (3DPW) domains.}
\label{table:domain_transfer}
\end{table}

\begin{figure*}
\centering
\setlength\tabcolsep{2pt}
\begin{tabular}{cccc|ccc}
 & \multicolumn{3}{c|}{InterHand2.6M} & \multicolumn{3}{c}{3DPW} \\
ViTPose++ & \raisebox{-0.5\height}{\includegraphics[height=0.36\columnwidth]{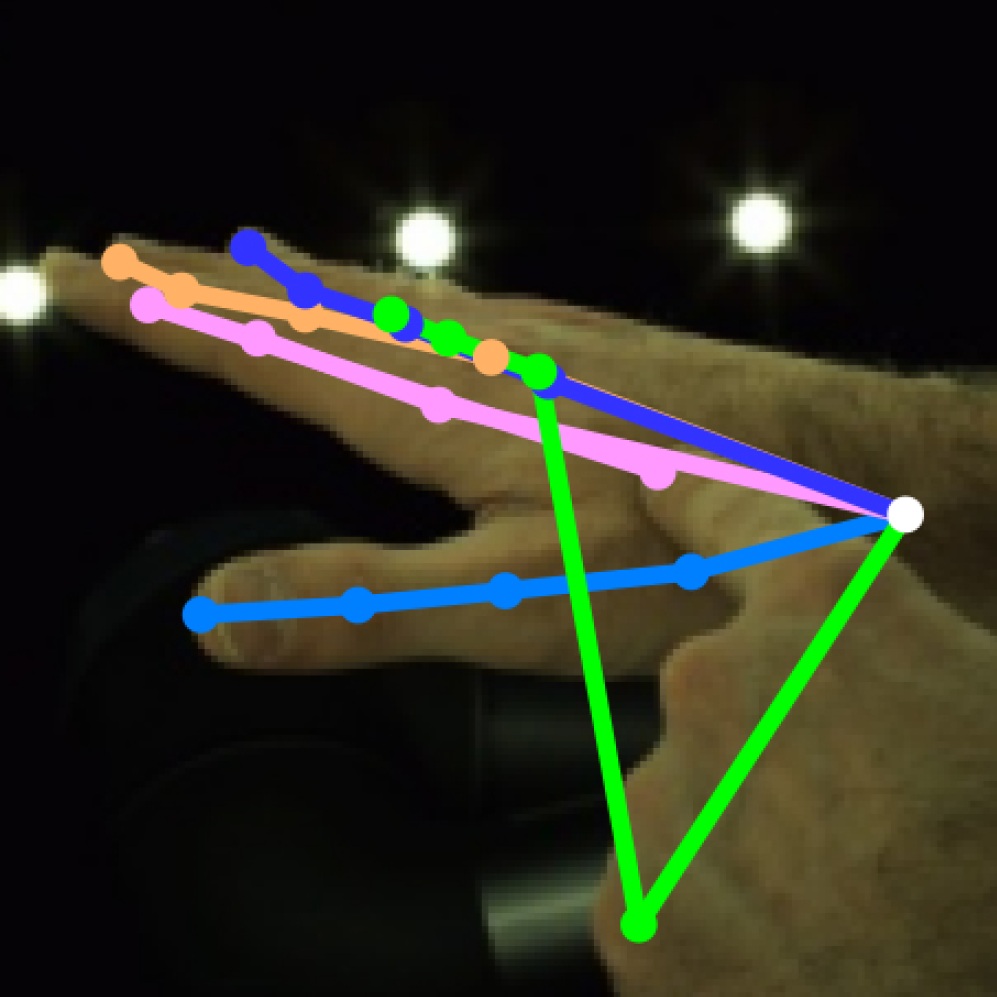}} & \raisebox{-0.5\height}{\includegraphics[height=0.36\columnwidth]{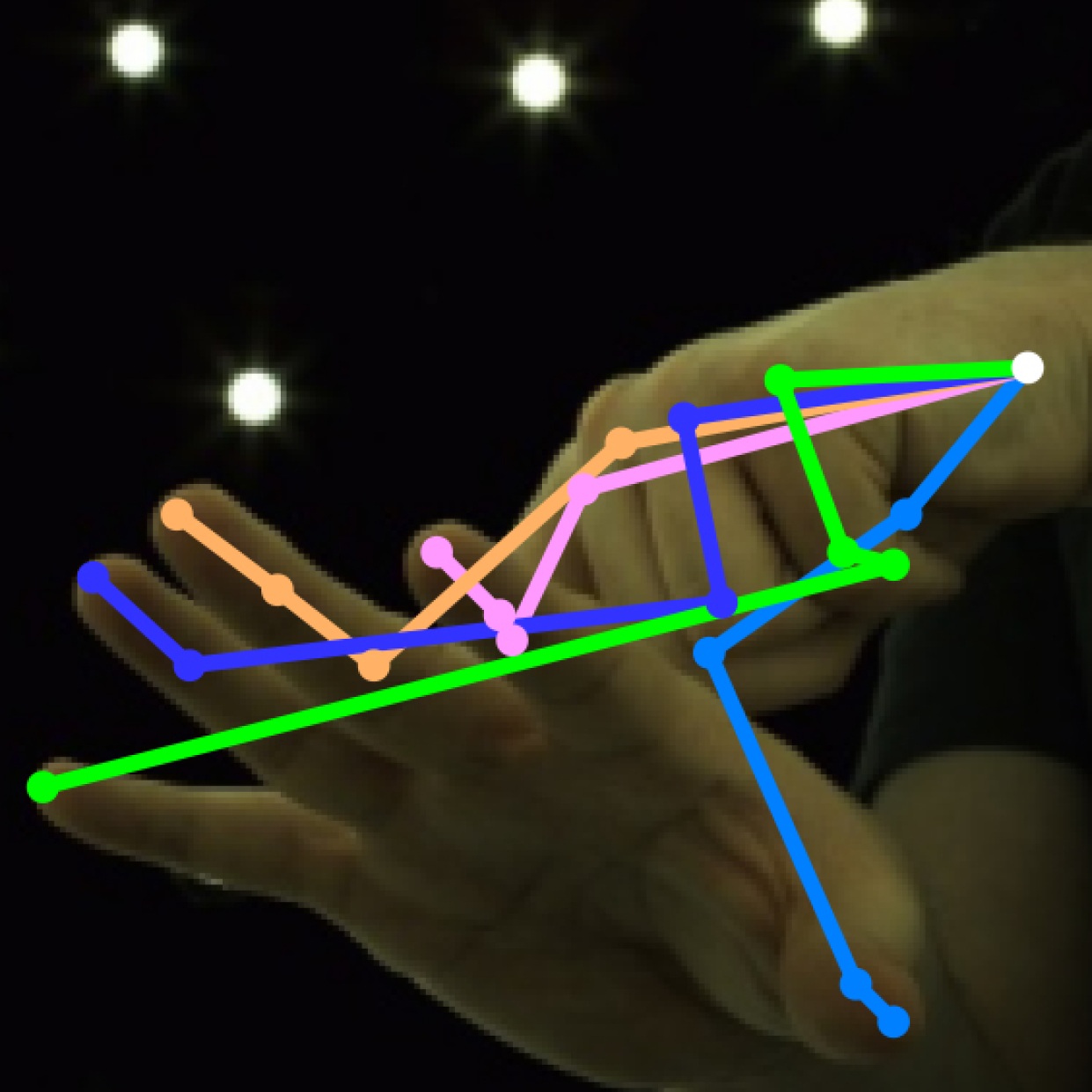}} & 
\raisebox{-0.5\height}{\includegraphics[height=0.36\columnwidth]{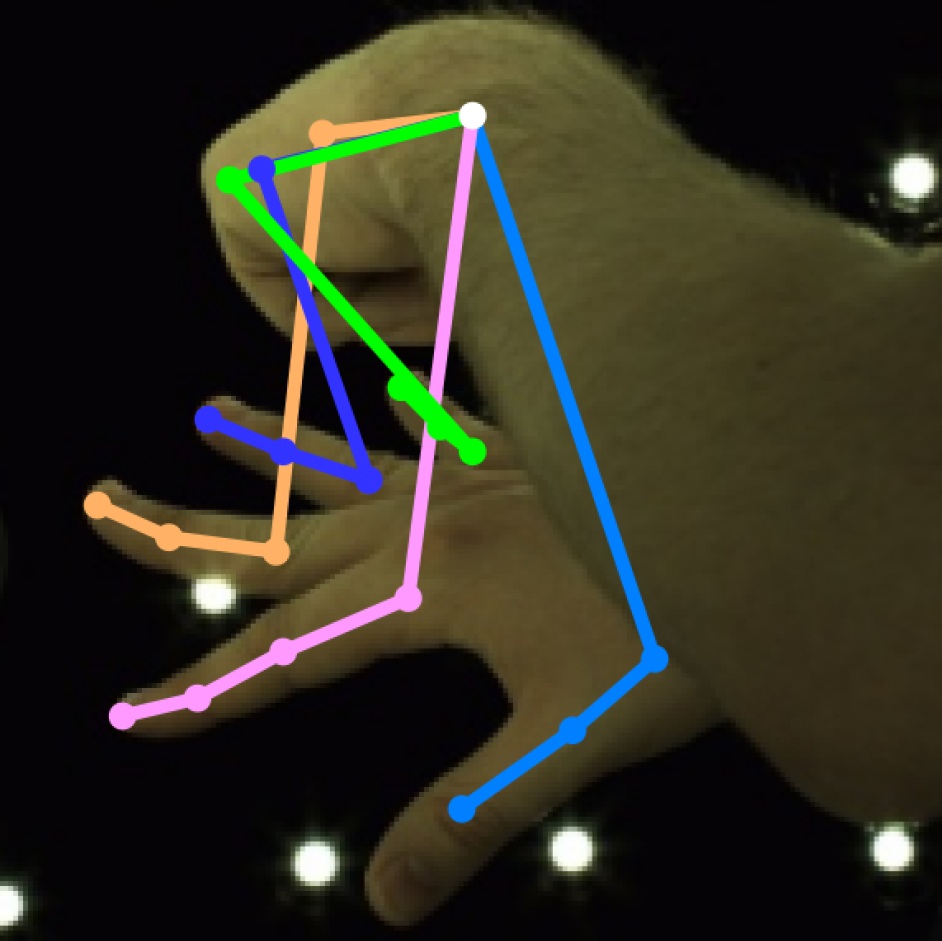}} &
\raisebox{-0.5\height}{\includegraphics[height=0.36\columnwidth]{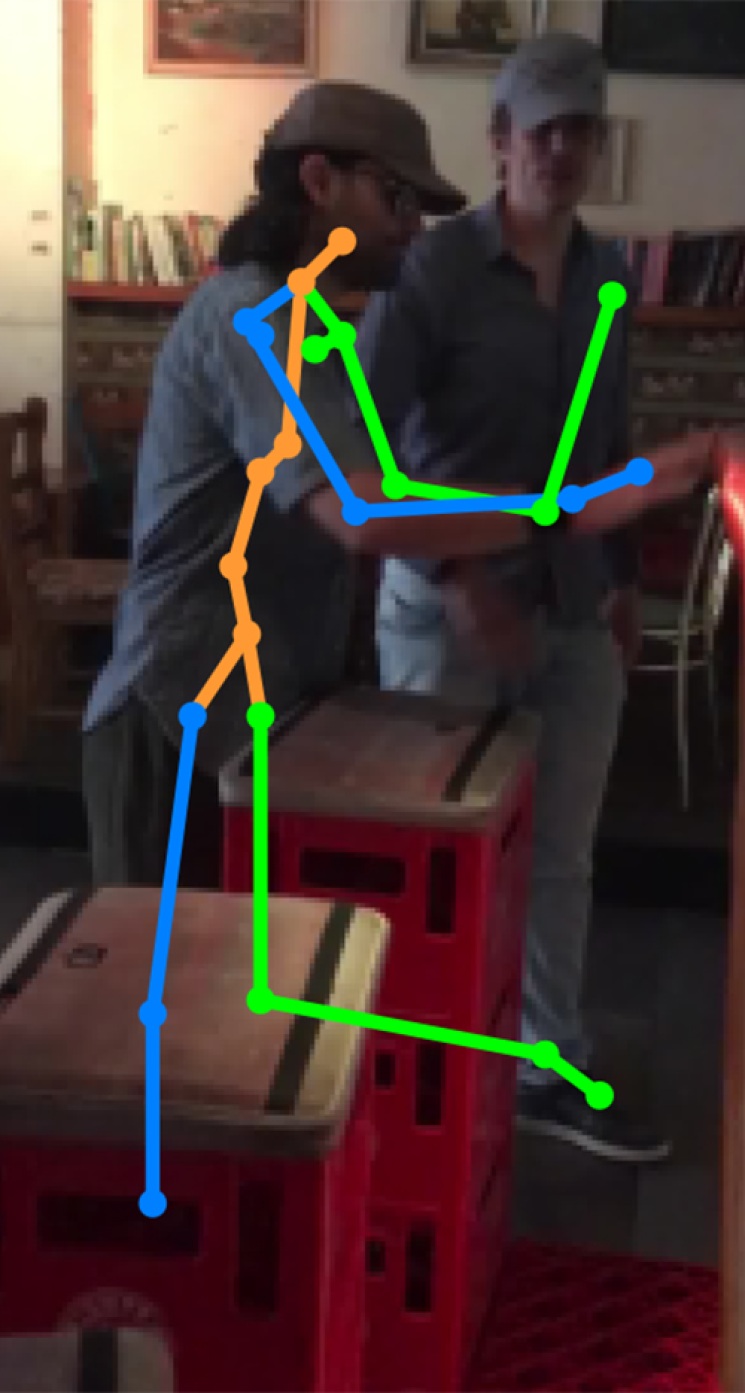}} & \raisebox{-0.5\height}{\includegraphics[height=0.36\columnwidth]{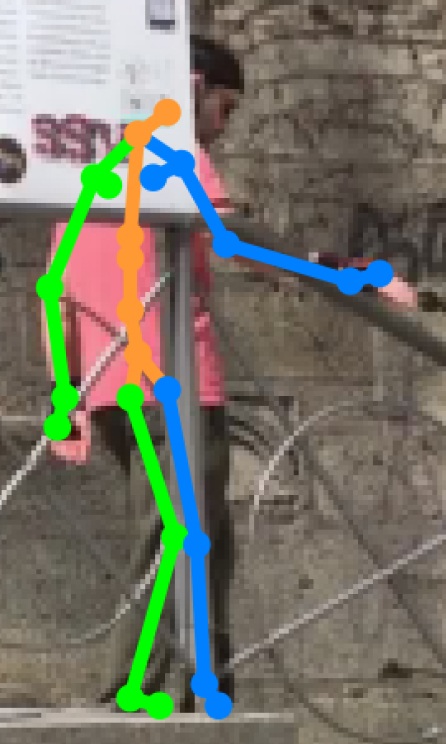}} &
\raisebox{-0.5\height}{\includegraphics[height=0.36\columnwidth]{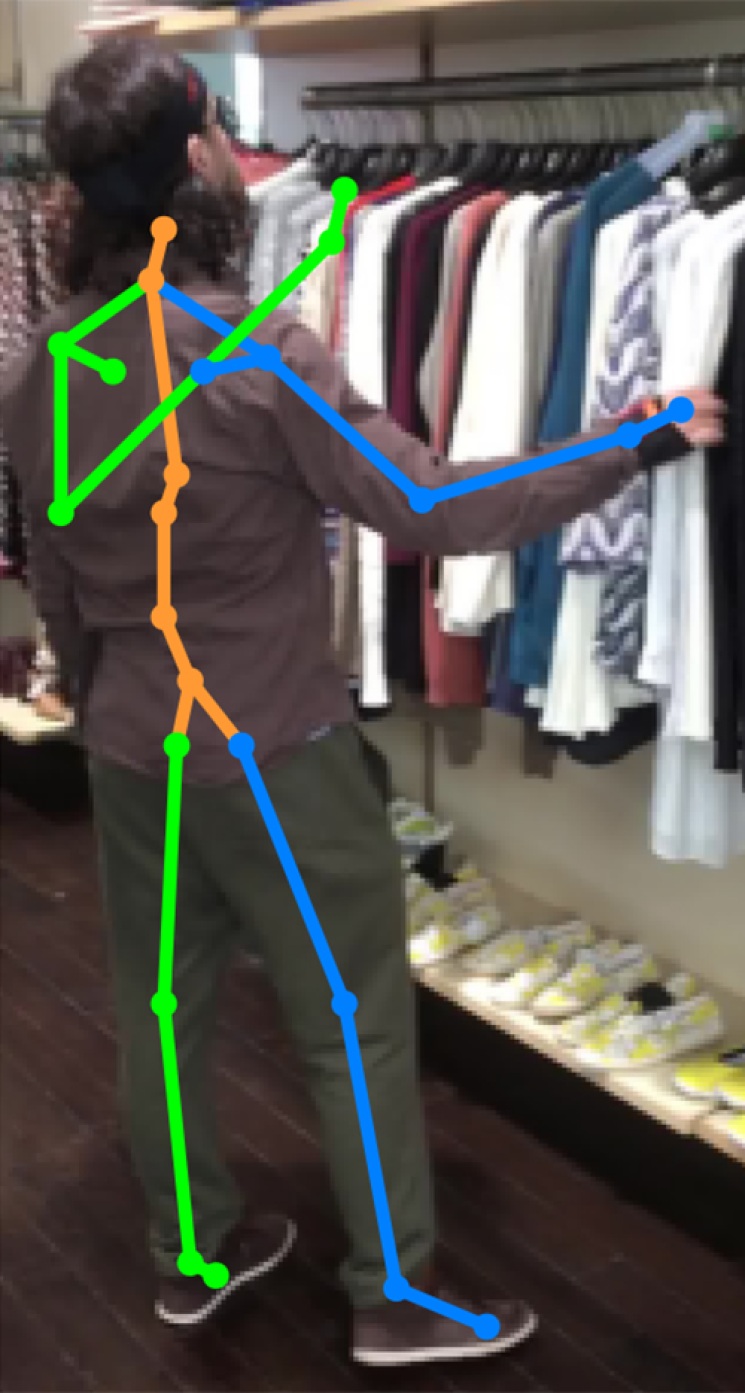}} \\
Ours & \raisebox{-0.5\height}{\includegraphics[height=0.36\columnwidth]{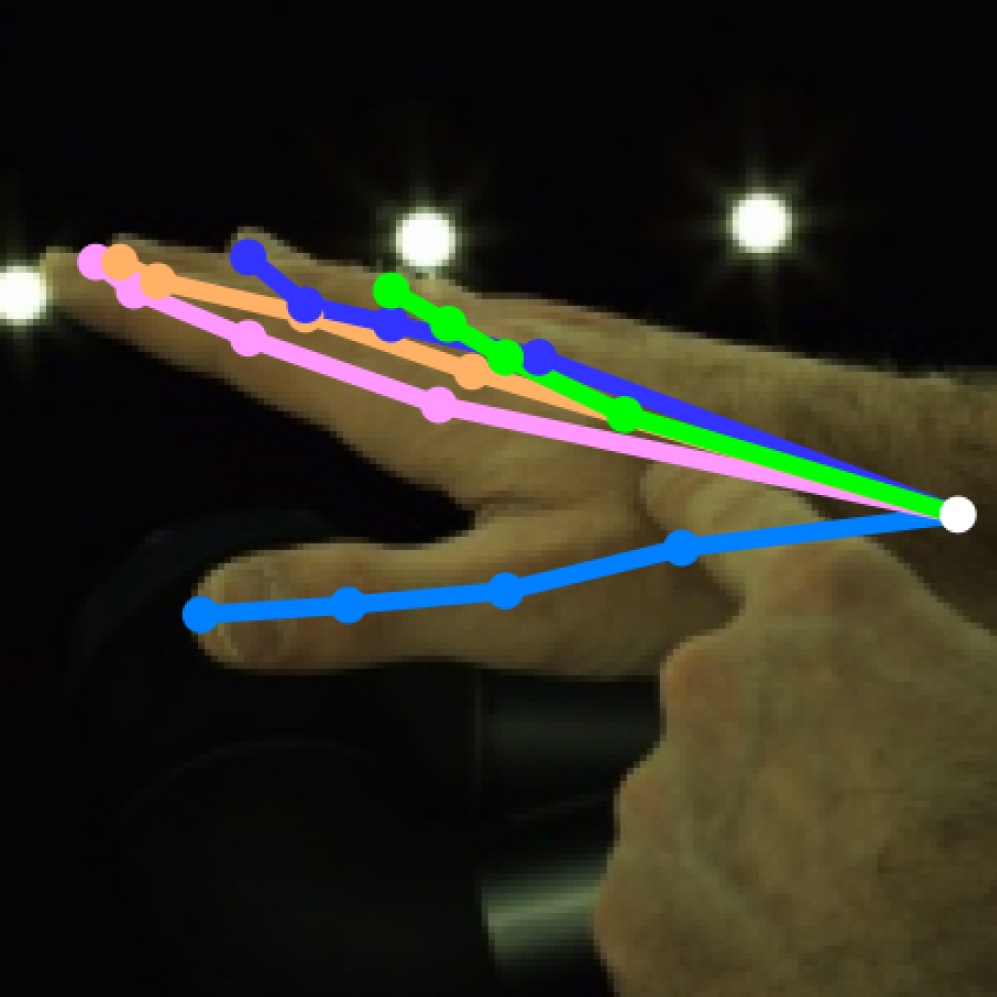}} & \raisebox{-0.5\height}{\includegraphics[height=0.36\columnwidth]{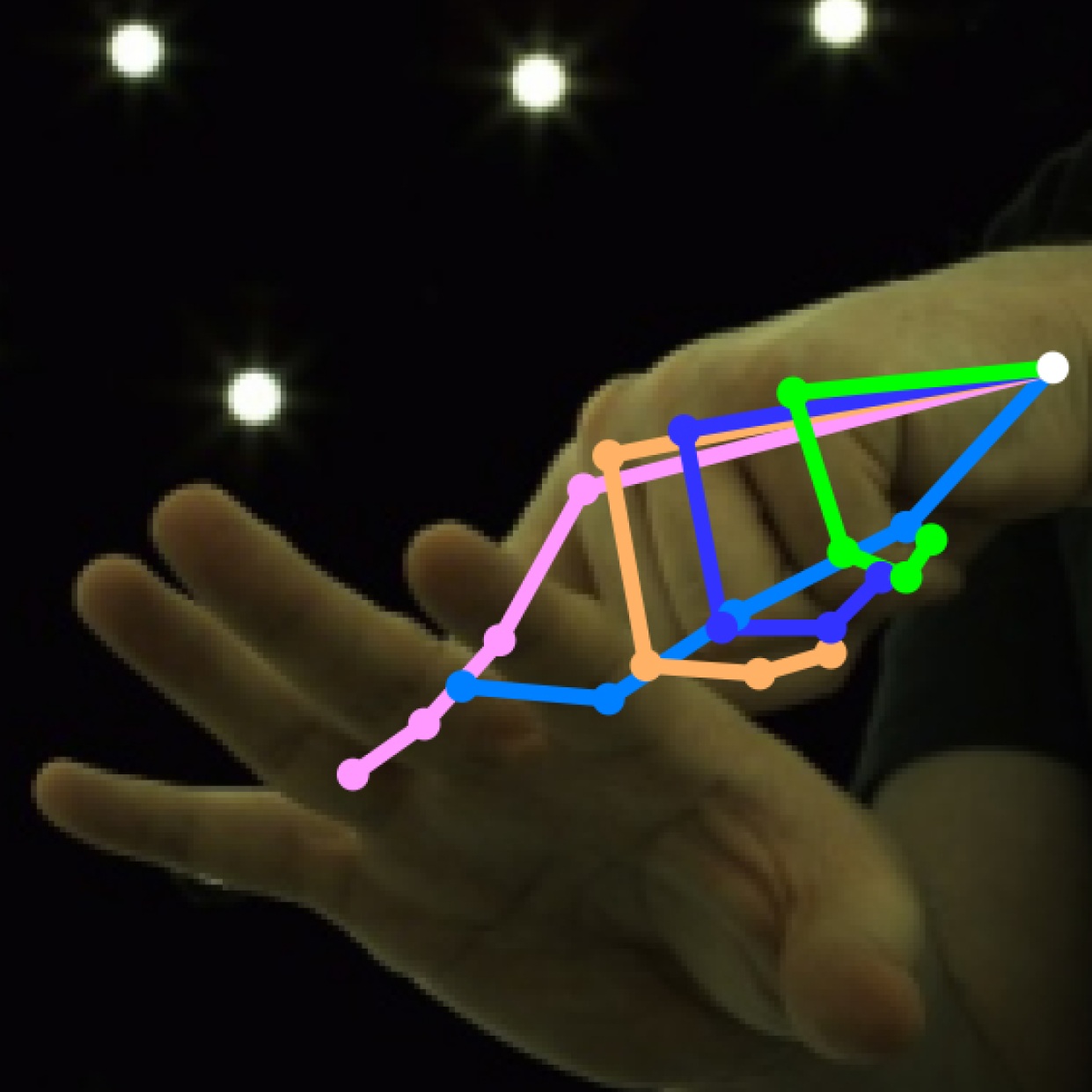}} & 
\raisebox{-0.5\height}{\includegraphics[height=0.36\columnwidth]{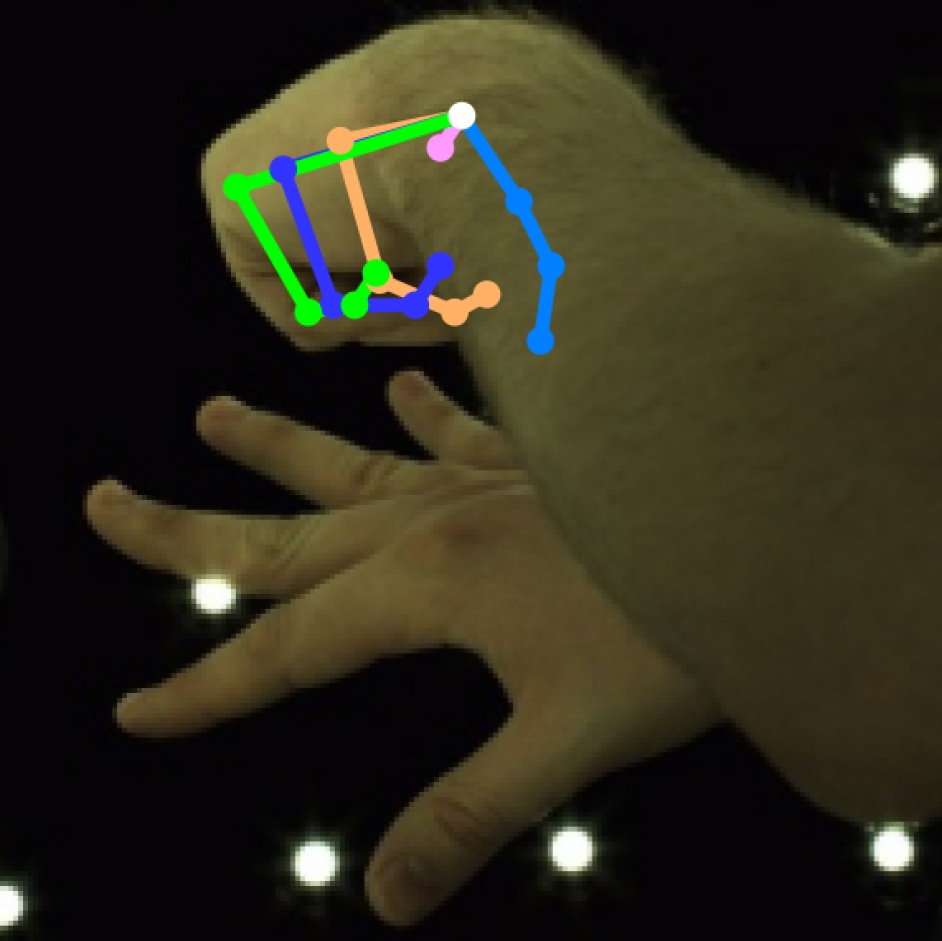}} &
\raisebox{-0.5\height}{\includegraphics[height=0.36\columnwidth]{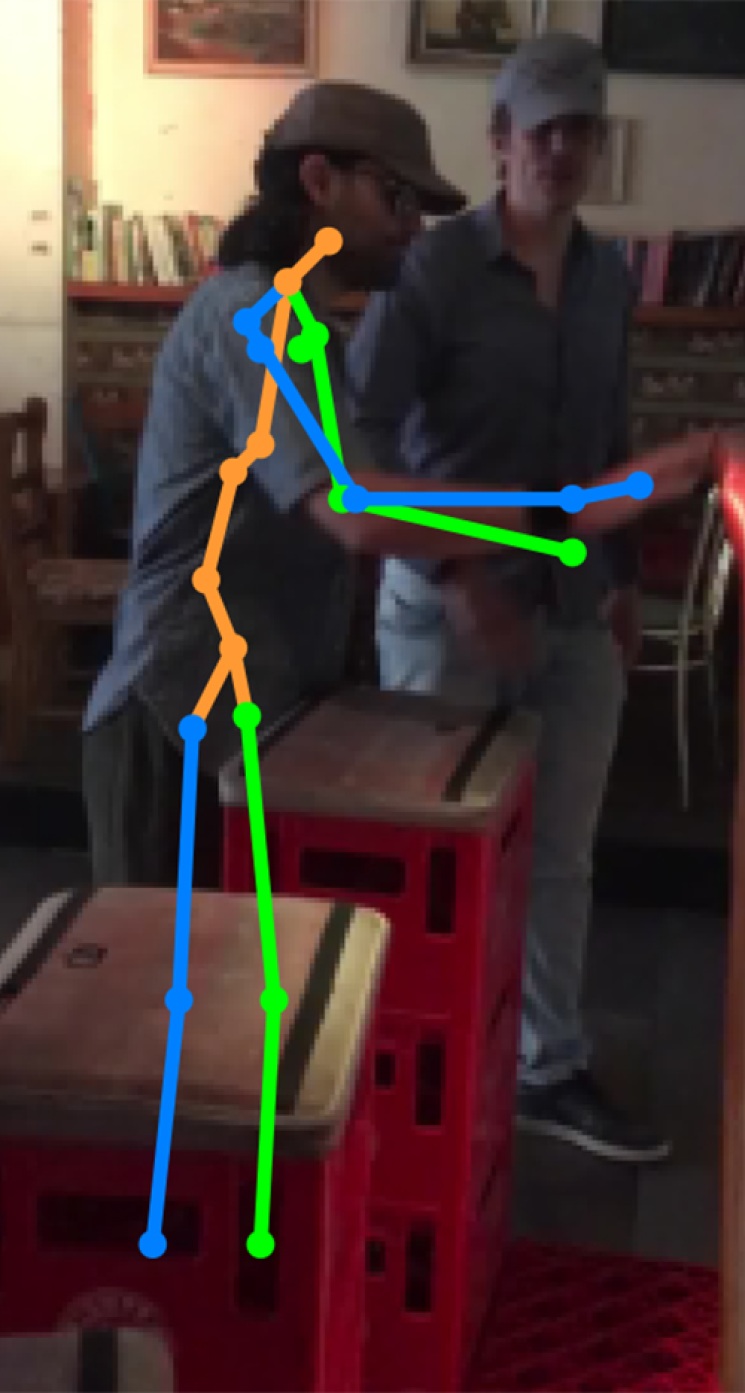}} & \raisebox{-0.5\height}{\includegraphics[height=0.36\columnwidth]{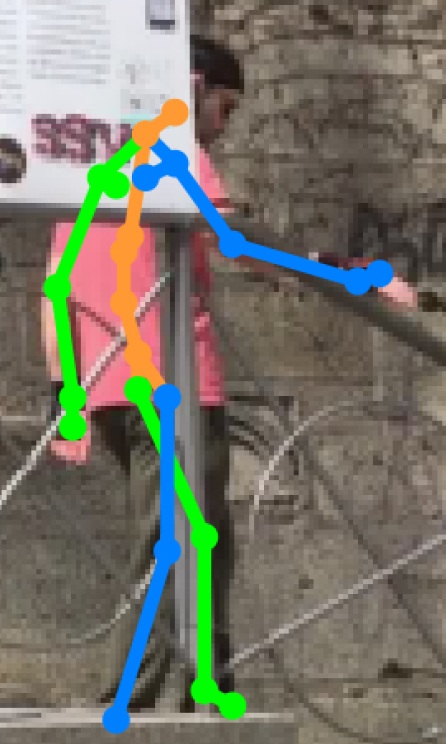}} &
\raisebox{-0.5\height}{\includegraphics[height=0.36\columnwidth]{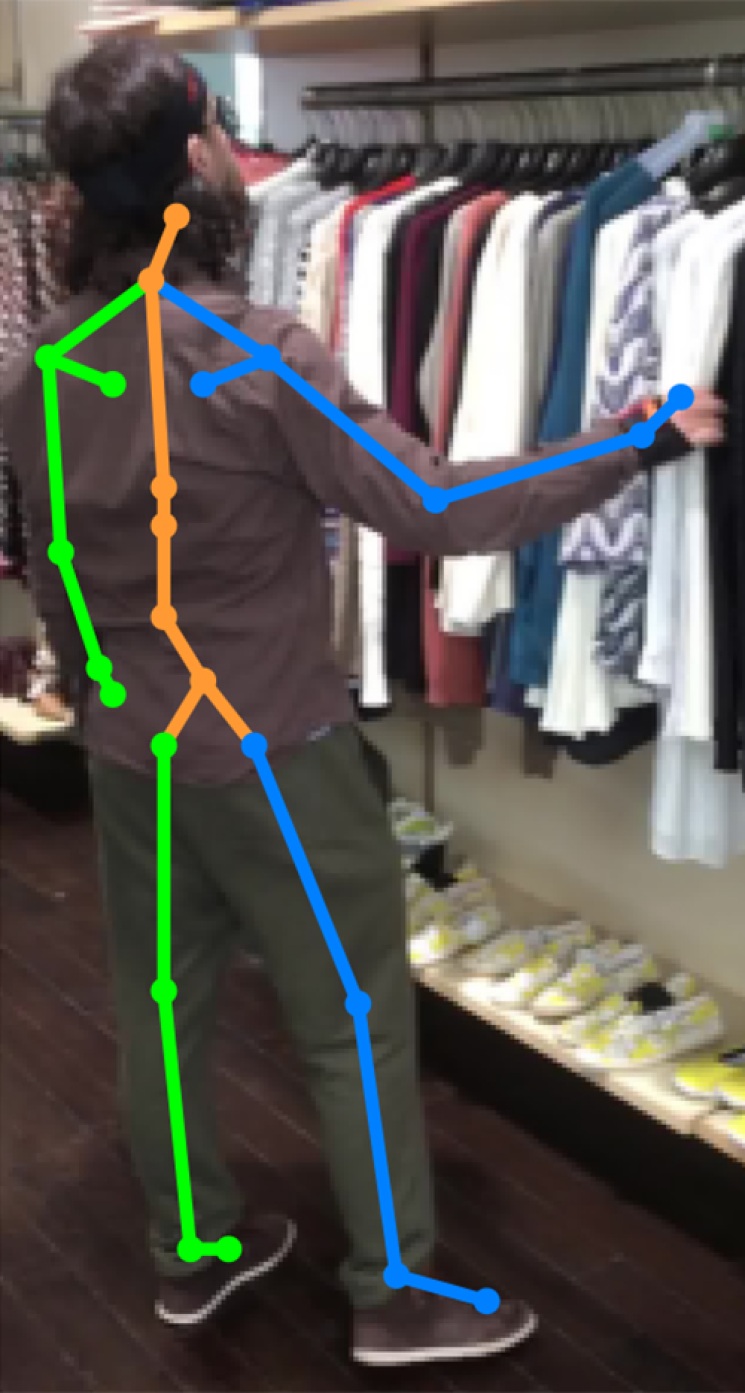}}\\
&(a) & (b) & (c) & (d) & (e) & (f) \\
\end{tabular}
\vspace{-0.15cm}
\caption{Comparative results on InterHand2.6M (a--c) and 3DPW (e--f).}
\label{fig:kpt_pred_transfer}
\end{figure*}

\begin{figure}
\centering
\footnotesize
\setlength\tabcolsep{1pt}
\begin{tabular}{ccccc}
& (a) & (b) & (c) & (d) \\
Input &\raisebox{-0.5\height}{\includegraphics[height=0.35\columnwidth]{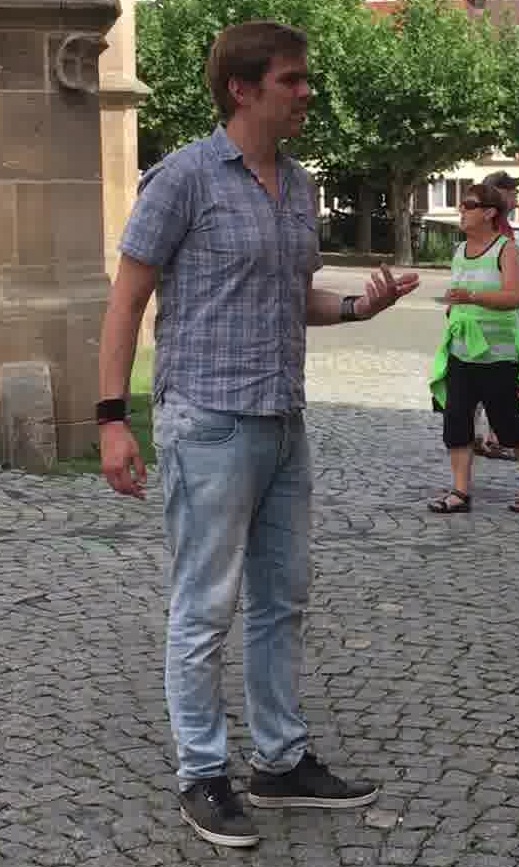}} & 
\raisebox{-0.5\height}{\includegraphics[height=0.35\columnwidth]{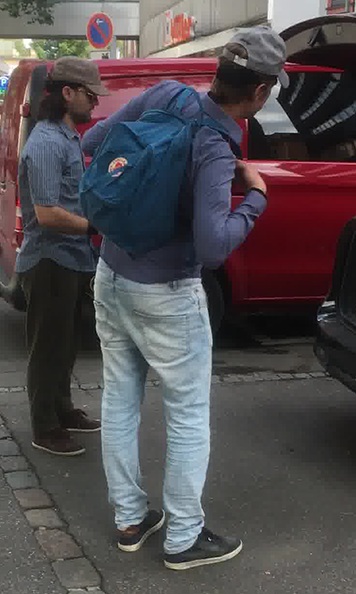}} &
\raisebox{-0.5\height}{\includegraphics[height=0.35\columnwidth]{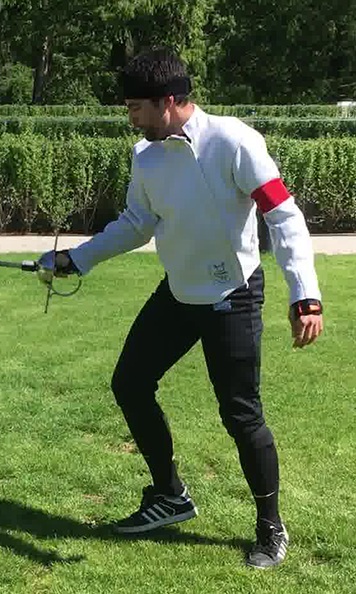}} &
\raisebox{-0.5\height}{\includegraphics[height=0.35\columnwidth]{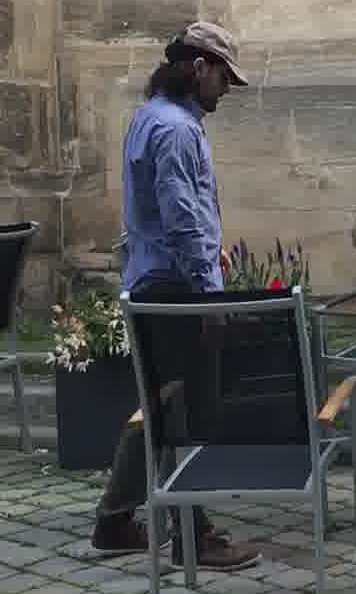}} \\
KITRO & \raisebox{-0.5\height}{\includegraphics[height=0.35\columnwidth]{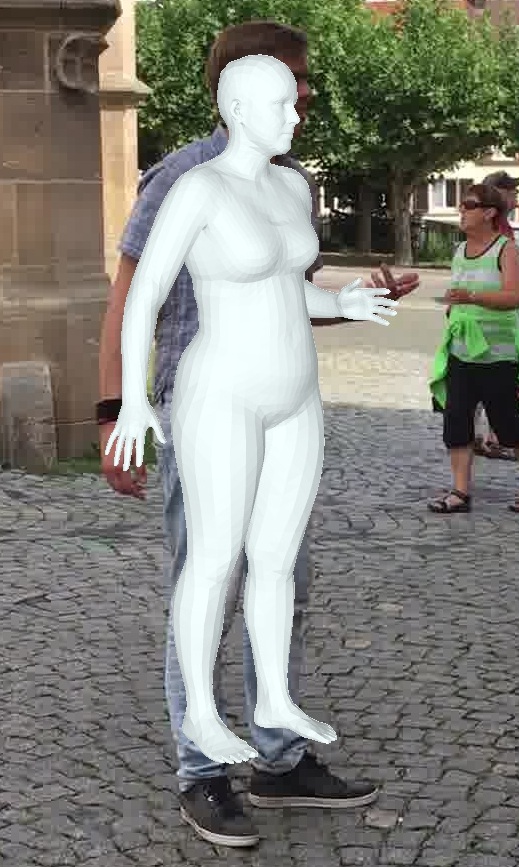}} & \raisebox{-0.5\height}{\includegraphics[height=0.35\columnwidth]{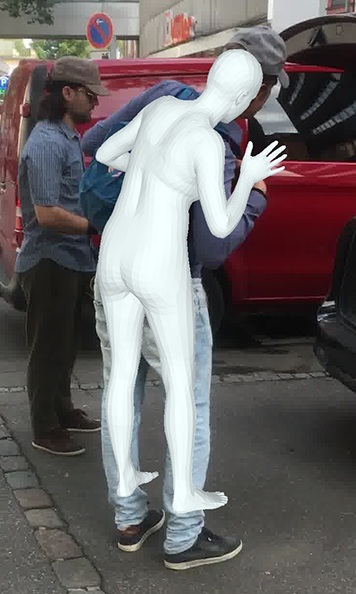}} &
\raisebox{-0.5\height}{\includegraphics[height=0.35\columnwidth]{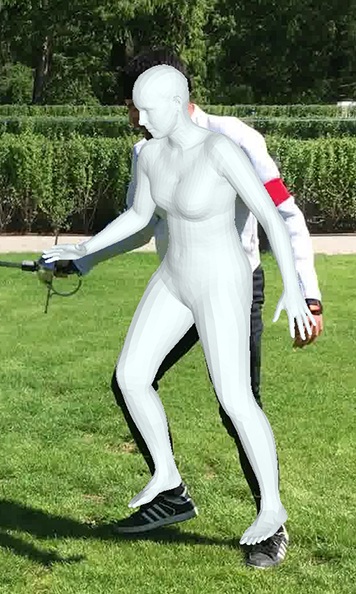}} &
\raisebox{-0.5\height}{\includegraphics[height=0.35\columnwidth]{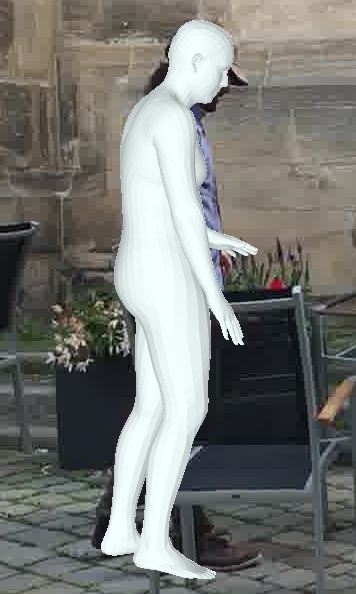}}\\
Ours & \raisebox{-0.5\height}{\includegraphics[height=0.35\columnwidth]{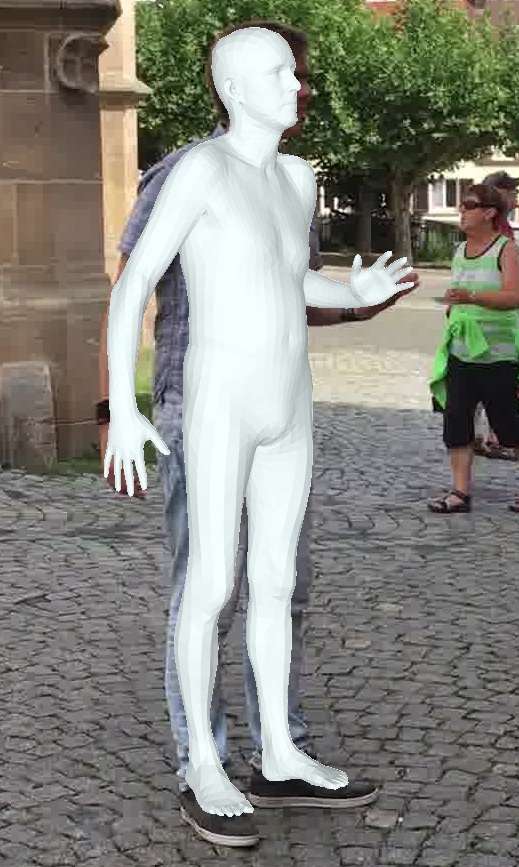}} &
\raisebox{-0.5\height}{\includegraphics[height=0.35\columnwidth]{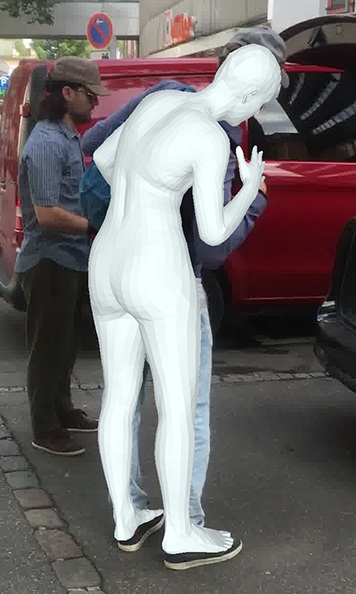}} &
\raisebox{-0.5\height}{\includegraphics[height=0.35\columnwidth]{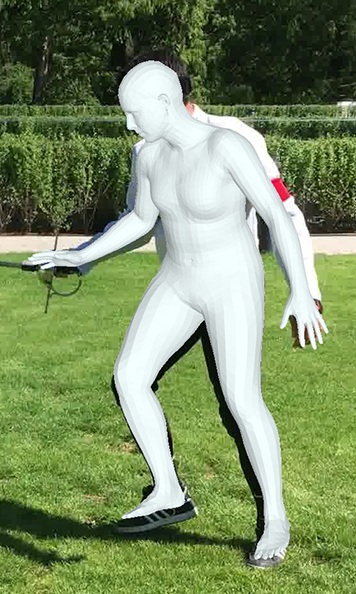}} &
\raisebox{-0.5\height}{\includegraphics[height=0.35\columnwidth]{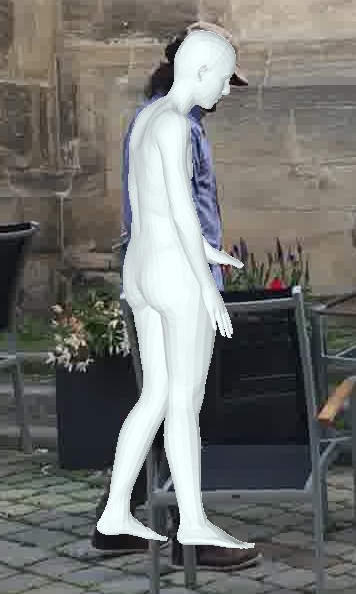}}\\
\end{tabular}
\caption{Examples of SMPL fitting using KITRO and with our 2D predictions.}
\label{fig:smpl_fit}
\end{figure}

\subsection{Transfer to hands and human shapes}
PoseBH has the capability to generalize its learned representations beyond the training dataset. We validate this through domain transfer experiments by fine-tuning the model separately on the InterHand2.6M and 3DPW datasets. InterHand2.6M is a two-hand pose estimation dataset containing 21 MANO keypoints~\cite{MANO_2017}, while 3DPW is an in-the-wild human shape estimation dataset with 24 SMPL keypoints~\cite{SMPL:2015}.

During fine-tuning, we freeze the embedding module and nonparametrically learn the prototypes for the target datasets. After convergence, we fine-tune the backbone and head weights alongside the prototype loss. Freezing the embedding module ensures that the previously learned prototypes and those of the target dataset remain aligned within the same embedding space, preserving the learned representations during transfer. Since these experiments do not involve skeleton heterogeneity across datasets, no semi-supervised learning challenges arise, and CSS loss is not applied. 

In~\cref{table:domain_transfer}, we report results on the InterHand2.6M and 3DPW test sets using the ViT-B backbone. Given its large dataset size ($\sim$2.6M instances), InterHand2.6M exhibits a smaller performance gap among different methods. Nonetheless, our method surpasses ViTPose++ by 0.3 PCK, 0.9 AUC, and 0.32 EPE, while also outperforming UniHCP by 0.1 PCK, 0.1 AUC, and 0.02 EPE. For the smaller 3DPW dataset, our method achieves a notable performance gain, surpassing the baseline by 1.9 AP and 1.9 AR, demonstrating strong generalization in  keypoint embeddings.

\subsection{Ablation studies}
In \cref{table:mdt_ablation}, we present the results of ablation experiments on the COCO (CO), AIC (AI), MPII (MP), AP-10K (AP), APT-36K (APT), and COCO-Wholebody (CW) validation sets using the ViT-Base backbone. The top row represents a multi-head baseline with a shared backbone and dataset-specific heads. This conventional multi-head approach exhibits relatively low performance, particularly on the AP-10K and APT-36K validation sets, due to substantial domain discrepancies. Our keypoint prototype method achieves an average score of 66.3, improving the baseline by 0.8. Additionally, incorporating cross-type self-supervision provides an additional average performance gain of 0.2. Collectively, these enhancements enable our final configuration to outperform the multi-head baseline by an average of 2.4 points across all six datasets.

\begin{table}[!t]
\centering
{
\setlength{\tabcolsep}{1mm}
\begin{tabular}{l|cccccc|c}
\toprule
Method & CO & AI & MP & AP & APT & CW & \textit{Avg.} \\ \midrule
Baseline & 77.0 & 31.6 & 93.1 & 75.3 & 74.8 & 57.1 & 68.2 \\ \midrule
+$\mathcal{L}_{\text{Proto}}$ & 77.0 & 31.8 & 93.0 & 76.4 & 86.4 & 57.5 & 70.4 \\
+$\mathcal{L}_{\text{CSS}}$ & \textbf{77.3} & \textbf{32.1} & \textbf{93.2} & \textbf{76.7} & \textbf{86.5} & \textbf{57.9} & \textbf{70.6} \\ 
\bottomrule
\end{tabular}
}%
\caption{Ablation study of the proposed method. $\mathcal{L}_{\text{Proto}}$: keypoint prototype learning. $\mathcal{L}_{\text{CSS}}$: cross-type self-supervision.
}
\label{table:mdt_ablation}
\end{table}

\subsection{Visual analysis}
\Cref{fig:kpt_pred} presents examples of human and animal pose estimation, demonstrating that our algorithm achieves more accurate keypoint localization under complex interactions and occlusions, as seen in the right-hand keypoints of the standing person (left) and the right front toe of the dog (right).

We further compare the pose estimation results on InterHand2.6M and 3DPW in \cref{fig:kpt_pred_transfer}. In (a), ViTPose++ incorrectly assigns a little finger joint to the opposite hand, whereas our method provides more accurate localization. In (b), ViTPose++ struggles with occlusion from the other hand, producing noisy predictions, while our method correctly separates them. In (c), due to severe self-occlusion, ViTPose++ misestimates occluded keypoints and assigns them to the opposite hand, while our method generalizes well in challenging self-occlusion scenarios. In (d), ViTPose++ produces an implausible foot and hand pose under heavy occlusion, while our method remains robust. In (e), ViTPose++ swaps the left and right legs, while our method correctly distinguishes them. In (f), background clutter causes ViTPose++ to produce noisy predictions for the left hand, while our method correctly identifies the location of the occluded left hand.

To further assess the effectiveness of our 2D SMPL keypoint predictions on the 3DPW test set, we employ a SMPL optimization technique. KITRO~\cite{KITRO_2024} optimizes SMPL parameters using 2D keypoint reprojection loss. For a comparative visual analysis, we replace KITRO's 2D keypoint predictions with ours. As shown in \Cref{fig:smpl_fit}(a) and (c), KITRO fails to align the mesh with the actual feet, whereas our method achieves more precise alignment. In (b), KITRO mislocalizes the right hand under partial occlusion, while our 2D predictions remain robust. In (d), KITRO exhibits misalignment in both the hands and feet, while replacing its 2D predictions with ours significantly improves the alignment of peripheral body parts.

In the supplemental material, we provide a visualization of the prototypes generated by our algorithm, illustrating how they effectively capture the diversity of representations within the embedding space.

\section{Conclusion}
This paper presents a new multi-dataset training (MDT) framework for pose estimation across diverse domains and datasets. Traditional approaches encounter challenges with label heterogeneity, dataset-specific parameters, and limited domain generalization. Our approach addresses the limitations with two main contributions. First, to resolve label heterogeneity and dataset-specific parameters, we employ nonparametric prototypes to unify diverse keypoint annotations across multiple datasets. Second, to address the limited multi-dataset supervision, we propose a cross-type self-supervision mechanism. The proposed method has demonstrated substantial improvements over existing MDT across a broad range of datasets, including human pose, animal pose, hand shape, and human shape estimation. 

\vspace{2mm}
\noindent\textbf{Limitations and future work.\;} 
A limitation of our approach is that it requires at least a few-shot sample to learn prototypes, making zero-shot inference for unseen skeletons infeasible. Future work should explore extending our method for zero-shot human pose estimation. Future work should also investigate adapting the approach to 3D domains \eg using axis-angle rotation representations, to facilitate representation learning on 3D manifolds.

\vspace{2mm}
\noindent\textbf{Acknowledgments.\;} This work was supported by the National Research Foundation of Korea (NRF) grant (No.~ 2021R1A2C2012195) and by the Institute of Information \& communications Technology Planning \& Evaluation (IITP) grants No.~RS-2019-II191906 (AI Graduate School Program, POSTECH), No.~RS-2020-II201336 (AI Graduate School Program, UNIST), and RS-2022-II220290, funded by the Korea government (MSIT).

{
    \small
    \bibliographystyle{ieeenat_fullname}
    \bibliography{main}
}

\newpage
\appendix
\section{Supplemental material}

\noindent In this supplementary material, we provide details on the keypoint embedding module~(\cref{s:keypointembeddingmodule}) and experimental settings~(\cref{s:experimentalsetup}), as well as additional experimental results~\cref{s:additionalresults}).

\section{Keypoint embedding module details}
\label{s:keypointembeddingmodule}
Our keypoint embedding module consists of two upsampling layers, one residual block, and two convolution layers. For the upsampling layers, we adopt the head design from ViTPose++~\cite{xu2023vitposepp}. The residual block combines a standard 3$\times$3 convolution with a skip connection, where we replace the ReLU activation with SiLU~\cite{elfwing2018sigmoid}. The final two convolution layers include a 3$\times$3 convolution, batch normalization, SiLU activation, and a 1$\times$1 output convolution. Although lightweight and simple, our embedding module effectively enhances multi-dataset training. We initialize the prototypes using a truncated normal distribution with a mean of 0, a standard deviation of 0.02, and a range of $[-2,2]$, followed by L2-normalization across the embedding dimension.

\section{Experimental setup}
\label{s:experimentalsetup}
\subsection{Hyperparameters}
We set the embedding dimension to $F=64$, the number of in-class prototypes to $M=3$, and the total number of keypoints to $J=214$. The output heatmap dimensions were set to a width of $W=48$ and a height of $H=64$.  Following~\cite{zhou2022rethinking}, we set $\kappa=0.05$ for obtaining the target $t_j$. The prototype update momentum $\lambda$ was set to 0.999. For the loss weights, we set $\alpha=3.33\times 10^{-6}$, $\beta=1.25\times 10^{-7}$, $\gamma=1.25\times 10^{-8}$, $\delta=0.01$, and $\zeta=0.001$. The impact of varying hyperparameter values (on the COCO validation set, measured in mean average precision; AP) is presented in~\cref{table:emb_dim,table:hyperparam,table:delta_zeta}, demonstrating robust performance across different hyperparameter settings. The final hyperparameter values are highlighted in bold.

\begin{table}[t]
\begin{center}
\begin{tabular}{r|c}
\toprule
$F$ & AP \\ \midrule
32 & 77.1 \\
\textbf{64}& 77.1 \\
128 & 77.1 \\
\bottomrule
\end{tabular}
\end{center}
\caption{Impact of varying the embedding dimension $F$ (mean AP on COCO validation set).}
\label{table:emb_dim}
\end{table}
\begin{table}[t]
\begin{center}
\begin{tabular}{c|c}
\toprule
$\alpha$ & AP \\ \midrule
$\mathbf{{3.33\times 10^{-6}}}$& 77.1 \\
$6.25\times 10^{-6}$ & 77.1 \\
$1.25\times 10^{-5}$ & 77.1 \\
\bottomrule
\end{tabular}
\begin{tabular}{cc|c}
\toprule
$\beta$ & $\gamma$ & AP \\ \midrule
$1.00\times 10^{-7}$ & $1.00\times 10^{-8}$ & 77.1 \\
$\mathbf{1.25\times 10^{-7}}$ & $\mathbf{1.25\times 10^{-8}}$ & 77.1 \\
$5.00\times 10^{-7}$ & $5.00\times 10^{-8}$ & 77.1 \\
$1.00\times 10^{-6}$ & $1.00\times 10^{-7}$ & 77.1 \\
\bottomrule
\end{tabular}
\end{center}
\caption{Impact of varying the loss weight values $\alpha$, $\beta$, and $\gamma$ (mean AP on COCO validation set).}
\label{table:hyperparam}
\end{table}

\begin{table}[t]
\begin{center}
\begin{tabular}{c|c}
\toprule
$\delta$ & AP \\ \midrule
$5.0\times 10^{-3}$ & 77.3\\
$\mathbf{1.0\times 10^{-2}}$ & 77.2\\
$5.0\times 10^{-2}$ & 77.3\\
\bottomrule
\end{tabular}
\begin{tabular}{c|c}
\toprule
$\zeta$ & AP \\ \midrule
$1.0\times 10^{-4}$ & 77.3\\
$\mathbf{1.0\times 10^{-3}}$ & 77.2\\
$1.0\times 10^{-2}$ & 77.3\\
\bottomrule
\end{tabular}
\end{center}
\caption{Impact of varying the loss weight values $\delta$ and $\zeta$ (mean AP on COCO validation set).}
\label{table:delta_zeta}
\end{table}

\subsection{APT-36K preprocessing}
For APT-36K~\cite{yang2022apt}, since official train, validation, and test splits are not provided, we partitioned the dataset using a 7:1:2 ratio following the guidelines of the original paper. This resulted in approximately 24,900 images and 37,000 instances for training, 3,600 images and 5,400 instances for validation, and 6,900 images and 10,700 instances in testing. To ensure that videos in the validation and test sets do not appear in the training set, each video is assigned to a single split.

\subsection{3DPW}
We use the processed annotations from ScoreHypo~\cite{xu2024scorehypo}. To obtain 2D keypoint annotations, we use the 2D projected 3D SMPL keypoints. We reformat the annotations to the COCO style and employ COCO-style evaluation metrics.

\subsection{Training.}
We trained our method on a system with four NVIDIA A100 GPUs or four NVIDIA A6000 GPUs using PyTorch 1.11 in an Ubuntu 20 environment. To enhance training efficiency, we also enabled automatic mixed precision with distributed training. We set the random seed value to 0 for all experiments to avoid randomness during training.

For learning rate scheduling, we start with an initial learning rate of 0.001 and reduce it by a factor of 0.1 at the 50th and 90th epochs. During the first 50 epochs, only the embedding module and prototypes are updated, while all other components remain frozen. At the start of the 50-th epoch, we set the backbone and the multi-heads to be trainable and freeze the prototypes. We then apply the $\mathcal{L}_{\text{CSS}}$ loss function (Eq. 8 in the main paper).

For transfer learning on InterHand2.6M, we follow ViTPose++ configurations. We train the model for 60 epochs with 5.0e-4 initial learning rate. In the case of transfer learning on 3DPW, we train the model for 30 epochs with 1.0e-4 initial learning rate. We train the prototypes for 30 epochs in InterHand2.6M, and 15 epochs in 3DPW.

\begin{table}[!t]
\centering
{
\setlength{\tabcolsep}{1mm}
\begin{tabular}{l|ccccccc}
\toprule
Method & AP & AP$^{50}$ & AP$^{75}$ & AP$^{M}$ & AP$^{L}$ & AR \\ \midrule
ViTPose++-B & 76.4 & 92.7 & 84.3 & 73.2 & 82.2 & 81.5 \\ 
ViTPose++-H & 78.5 & \textbf{93.4} & \textbf{86.2} & \textbf{75.3} & \textbf{84.4} & 83.4 \\ \midrule
\rowcolor[gray]{.8}
Ours-B & 76.6 & 92.6 & 84.4 & 73.4 & 82.4 & 81.7 \\
\rowcolor[gray]{.8}
Ours-H & \textbf{78.6} & 93.3 & \textbf{86.2} & \textbf{75.3} & \textbf{84.4} & \textbf{83.5} \\
\bottomrule
\end{tabular}
}%
\caption{COCO test-dev evaluation results.}
\label{table:supp_coco_test}
\end{table}

\begin{table}[!t]
\centering
{
\setlength{\tabcolsep}{1mm}
\begin{tabular}{l|cccc}
\toprule
Method & Val & Test & Val (occ) & Test (occ) \\ \midrule
ViTPose++-B & 81.1 & 82.0 & 64.0 & 64.1 \\ 
ViTPose++-H & 85.7 & 86.8 & 72.6 & 72.9 \\ \midrule
\rowcolor[gray]{.8}
Ours-B & 82.2 & 83.1 & 66.3 & 66.2 \\
\rowcolor[gray]{.8}
Ours-H & \textbf{86.0} & \textbf{87.0} & \textbf{73.2} & \textbf{73.7} \\
\bottomrule
\end{tabular}
}%
\caption{OCHuman evaluation results (measured in mean average precision; AP). Ground-truth bounding boxes were used for cropping.
}
\label{table:supp_ochuman}
\end{table}

\begin{figure*}[t]
\centering
\addtolength{\tabcolsep}{-0.4em}
\small
\begin{tabular}{cccccc}
 & Input & COCO & AIC & MPII & COCO-W \\
ViTPose++ & \raisebox{-0.5\height}{\includegraphics[height=0.5\columnwidth]{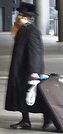}} & \raisebox{-0.5\height}{\includegraphics[height=0.5\columnwidth]{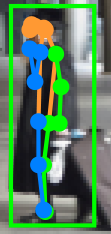}} & \raisebox{-0.5\height}{\includegraphics[height=0.5\columnwidth]{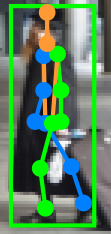}} & \raisebox{-0.5\height}{\includegraphics[height=0.5\columnwidth]{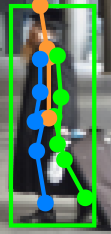}} & \raisebox{-0.5\height}{\includegraphics[height=0.5\columnwidth]{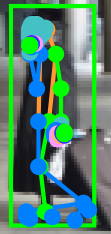}} \\
Ours & \raisebox{-0.5\height}{\includegraphics[height=0.5\columnwidth]{images_suppl/coco_000000012280/000000012280.png}} & \raisebox{-0.5\height}{\includegraphics[height=0.5\columnwidth]{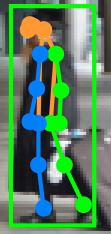}} & \raisebox{-0.5\height}{\includegraphics[height=0.5\columnwidth]{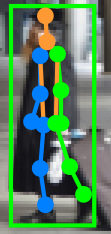}} & \raisebox{-0.5\height}{\includegraphics[height=0.5\columnwidth]{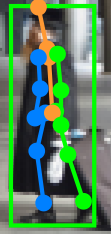}} & \raisebox{-0.5\height}{\includegraphics[height=0.5\columnwidth]{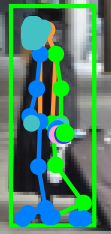}} \\
\vspace{2mm}\\
 & Input & COCO & AIC & MPII & COCO-W \\
ViTPose++ & \raisebox{-0.5\height}{\includegraphics[width=0.175\textwidth]{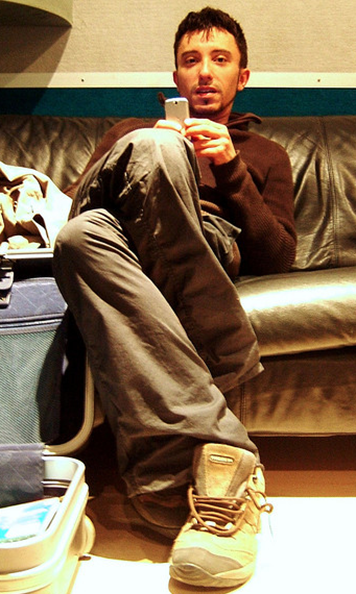}} & \raisebox{-0.5\height}{\includegraphics[width=0.175\textwidth]{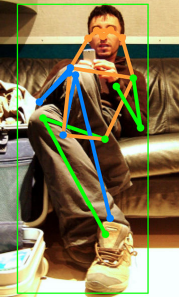}} & \raisebox{-0.5\height}{\includegraphics[width=0.175\textwidth]{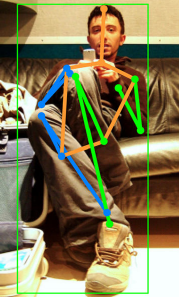}} & \raisebox{-0.5\height}{\includegraphics[width=0.175\textwidth]{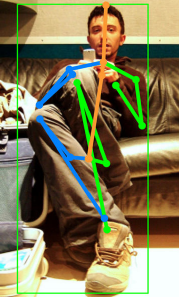}} & \raisebox{-0.5\height}{\includegraphics[width=0.175\textwidth]{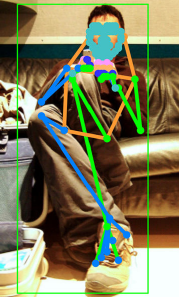}} \\
Ours & \raisebox{-0.5\height}{\includegraphics[width=0.175\textwidth]{images_suppl/coco_000000560911/000000560911.png}} & \raisebox{-0.5\height}{\includegraphics[width=0.175\textwidth]{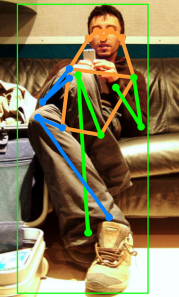}} & \raisebox{-0.5\height}{\includegraphics[width=0.175\textwidth]{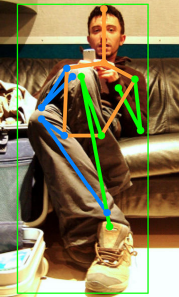}} & \raisebox{-0.5\height}{\includegraphics[width=0.175\textwidth]{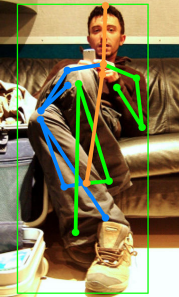}} & \raisebox{-0.5\height}{\includegraphics[width=0.175\textwidth]{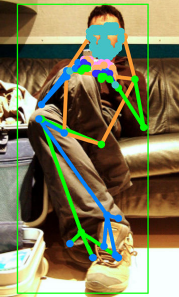}} \\
\end{tabular}
\caption{Pose estimation examples comparing ViTPose++ and our method.}
\label{fig:human_vis_suppl}
\end{figure*}

\begin{figure*}[t]
\centering
\begin{tabular}{ccc|cc}
 & AP-10K & APT-36K & AP-10K & APT-36K \\
ViTPose++ & \raisebox{-0.5\height}{\includegraphics[height=0.5\columnwidth]{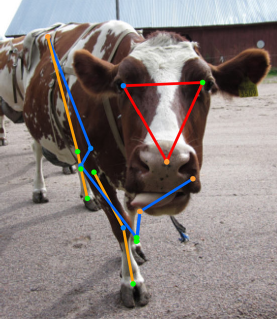}} & \raisebox{-0.5\height}{\includegraphics[height=0.5\columnwidth]{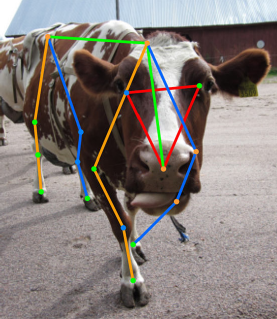}} & \raisebox{-0.5\height}{\includegraphics[height=0.5\columnwidth]{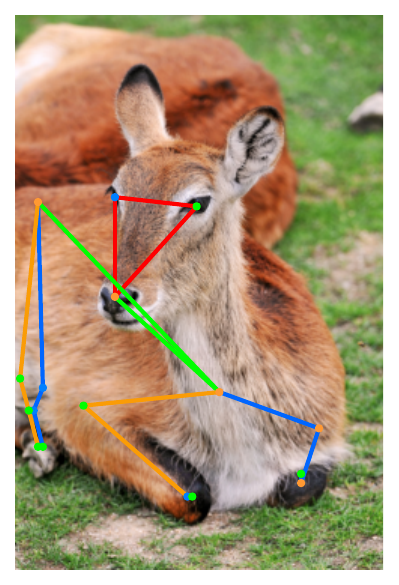}} & \raisebox{-0.5\height}{\includegraphics[height=0.5\columnwidth]{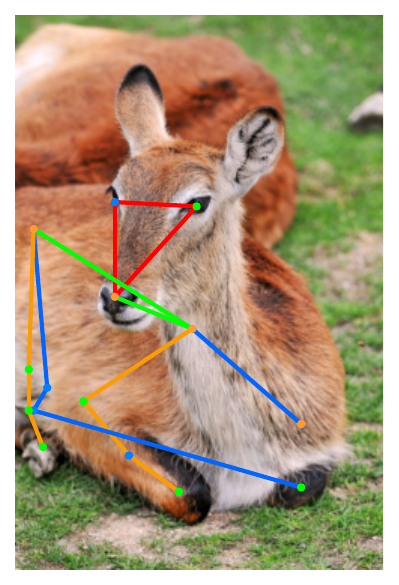}}\\
Ours & \raisebox{-0.5\height}{\includegraphics[height=0.5\columnwidth]{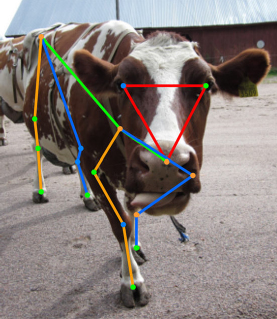}} & \raisebox{-0.5\height}{\includegraphics[height=0.5\columnwidth]{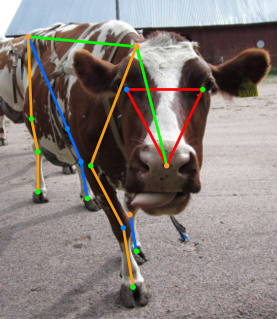}} & \raisebox{-0.5\height}{\includegraphics[height=0.5\columnwidth]{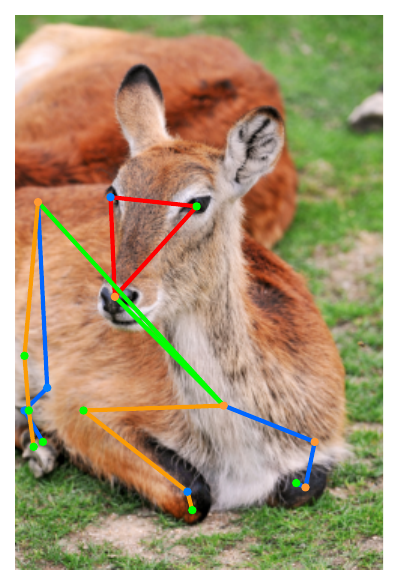}} & \raisebox{-0.5\height}{\includegraphics[height=0.5\columnwidth]{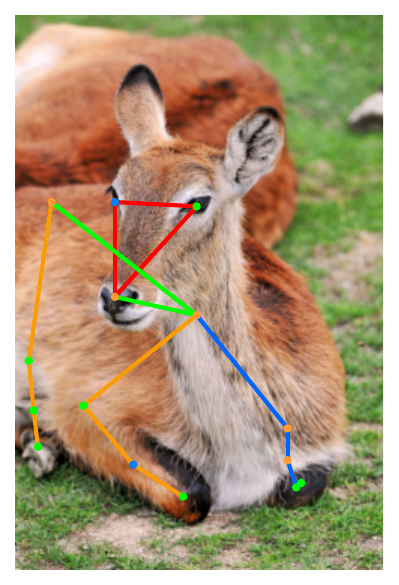}}\\
\end{tabular}
\caption{Pose estimation examples for animals using ViTPose++ and our method.}
\label{fig:animal_vis_suppl_1}
\end{figure*}

\begin{figure*}[t]
\centering
\footnotesize
\setlength\tabcolsep{1pt}
\begin{tabular}{ccccc}
\raisebox{-0.5\height}{\includegraphics[width=0.19\textwidth]{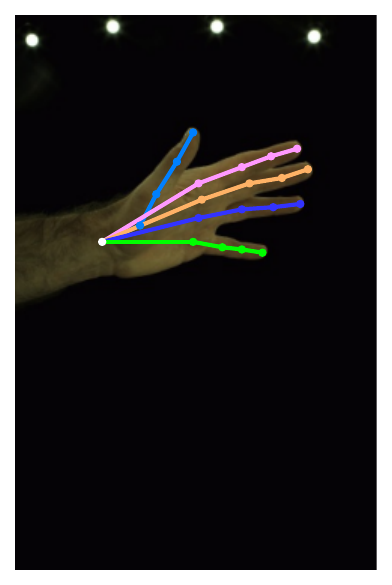}} & \raisebox{-0.5\height}{\includegraphics[width=0.19\textwidth]{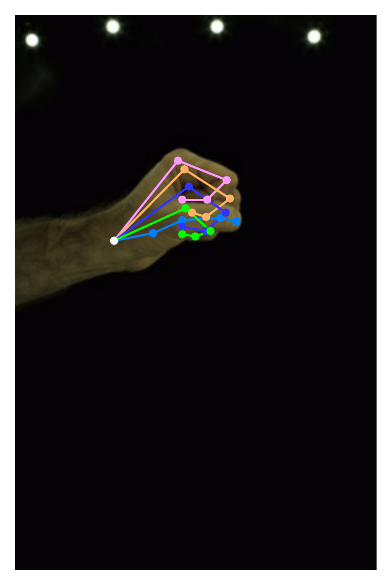}} & \raisebox{-0.5\height}{\includegraphics[width=0.19\textwidth]{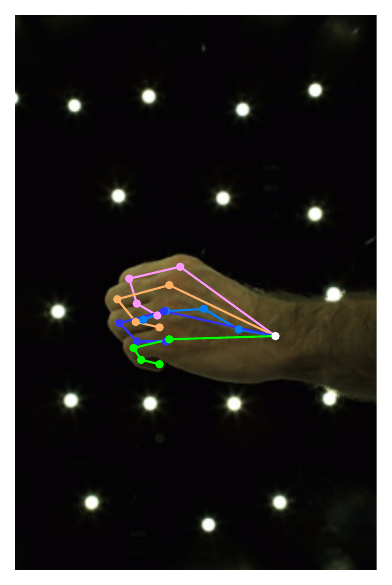}} & \raisebox{-0.5\height}{\includegraphics[width=0.19\textwidth]{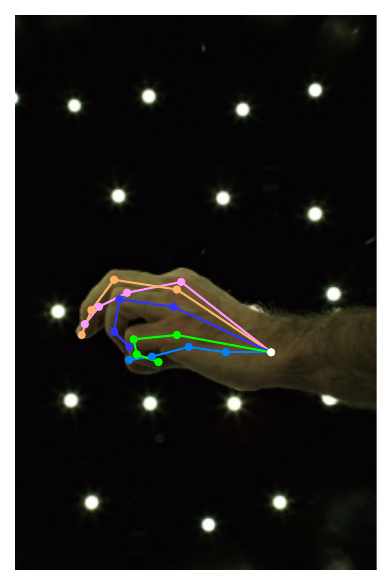}} & \raisebox{-0.5\height}{\includegraphics[width=0.19\textwidth]{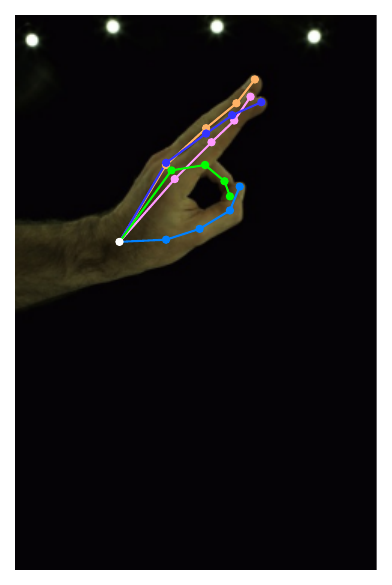}}\\
\end{tabular}
\caption{Pose estimation examples from our algorithm on the InterHand2.6M dataset.}
\label{fig:interhand_suppl}
\end{figure*}

\begin{figure*}[t]
\centering
\footnotesize
\setlength\tabcolsep{1pt}
\begin{tabular}{ccccc}
\raisebox{-0.5\height}{\includegraphics[width=0.19\textwidth]{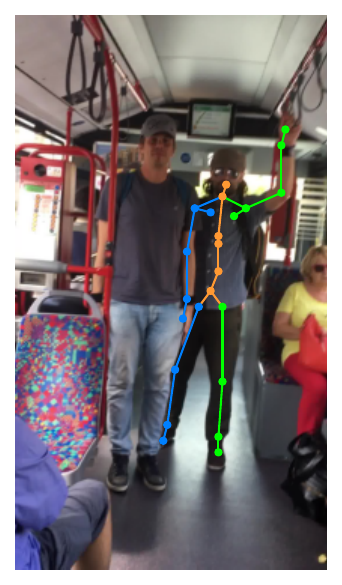}} & \raisebox{-0.5\height}{\includegraphics[width=0.19\textwidth]{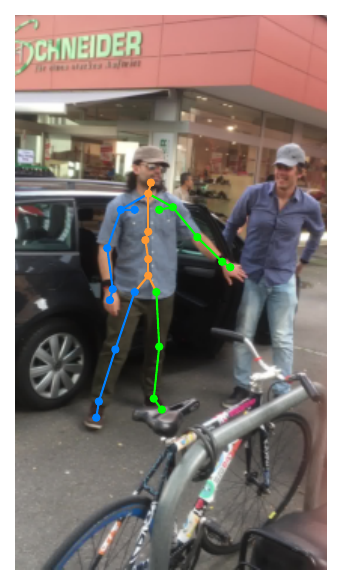}} & \raisebox{-0.5\height}{\includegraphics[width=0.19\textwidth]{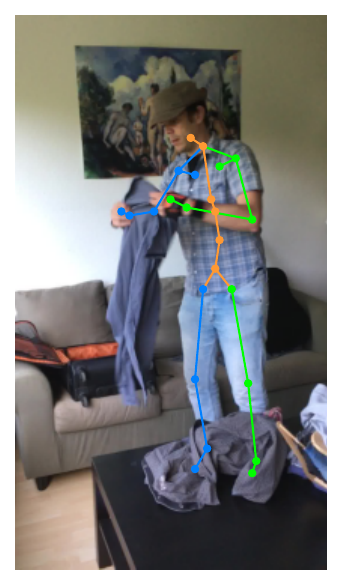}} & \raisebox{-0.5\height}{\includegraphics[width=0.19\textwidth]{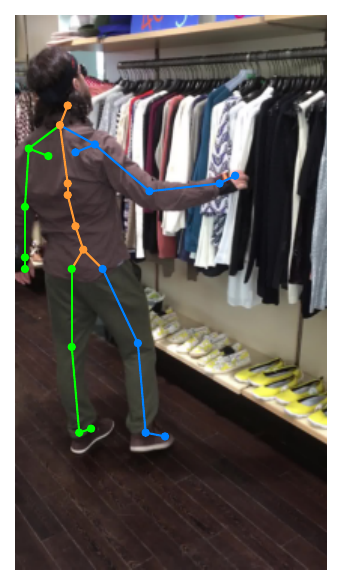}} & \raisebox{-0.5\height}{\includegraphics[width=0.19\textwidth]{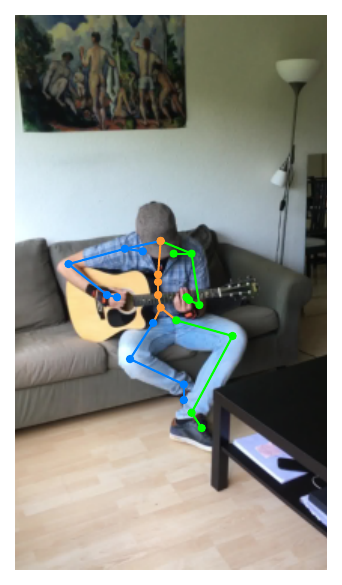}}\\
\end{tabular}
\caption{Pose estimation examples from our algorithm on the 3DPW dataset.}
\label{fig:3dpw_suppl}
\end{figure*}

\begin{table}[!t]
\centering
\begin{tabular}{l|c}
\toprule
Method & Average score \\ \midrule
ViTPose++ & 68.2 \\
+UniDet & 68.9 \\ \midrule
\rowcolor[gray]{.8}
Ours & \textbf{70.6} \\
\bottomrule
\end{tabular}
\caption{Performance of different MDT methods (average over six datasets).}
\label{table:supp_label_agg}
\end{table}

\begin{table}[!t]
\centering
\begin{tabular}{l|cc}
\toprule
Method & InterHand & 3DPW \\ \midrule
ViTPose++ & 86.2 & 81.7 \\
Trn. scratch & 86.1 & 56.8 \\ 
AIC-trained & 86.3 & 81.5 \\ \midrule
\rowcolor[gray]{.8}
Ours & \textbf{87.1} & \textbf{83.6} \\
\bottomrule
\end{tabular}
\caption{
Performance of domain transfer methods on the InterHand2.6M (AUC) and 3DPW (AP) datasets.
}
\label{table:supp_scratch}
\end{table}

\begin{figure*}[t]
\centerline{\includegraphics[width=0.7\textwidth]{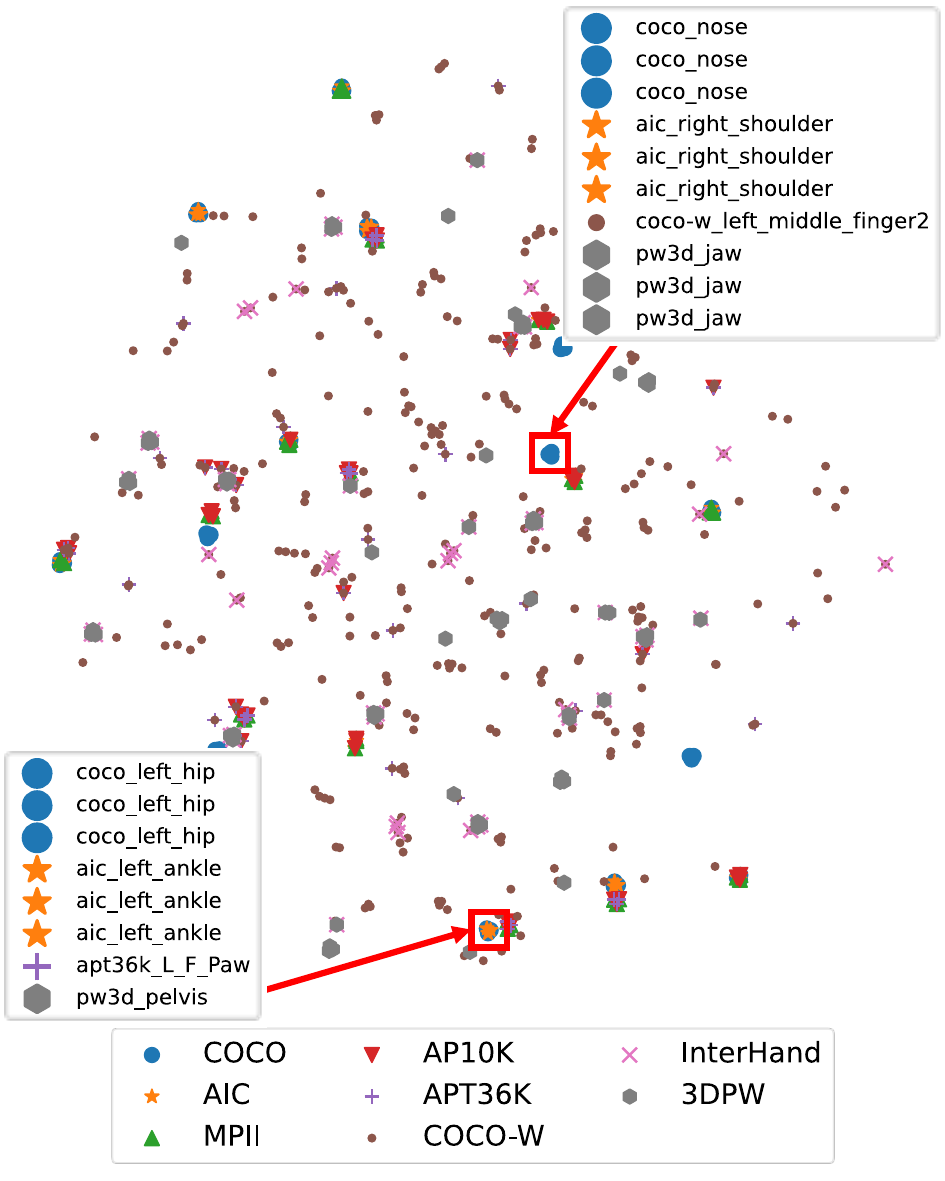}}
\caption{t-SNE visualization of the prototypes. Best viewed when zoom-in.}
\label{fig:emb_vis_suppl}
\end{figure*}

\begin{figure*}
\centering
\setlength\tabcolsep{1pt}
\begin{tabular}{cc|cc}
(a) & (b) & (c) & (d) \\
\raisebox{-0.5\height}{\includegraphics[height=0.5\columnwidth]{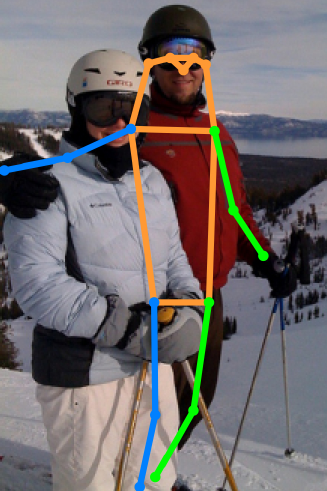}} &
\raisebox{-0.5\height}{\includegraphics[height=0.5\columnwidth]{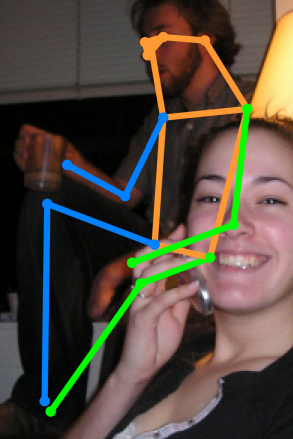}} & 
\raisebox{-0.5\height}{\includegraphics[height=0.5\columnwidth]{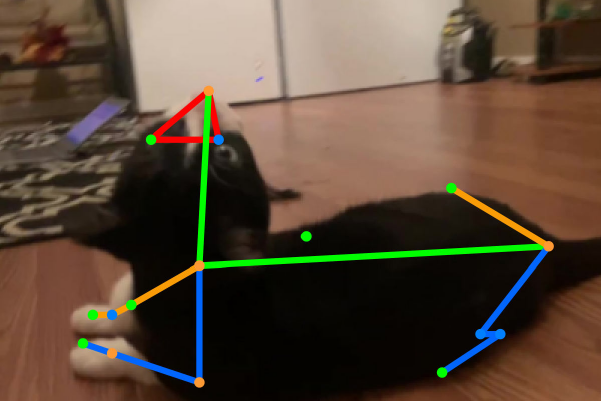}}&
\raisebox{-0.5\height}{\includegraphics[height=0.5\columnwidth]{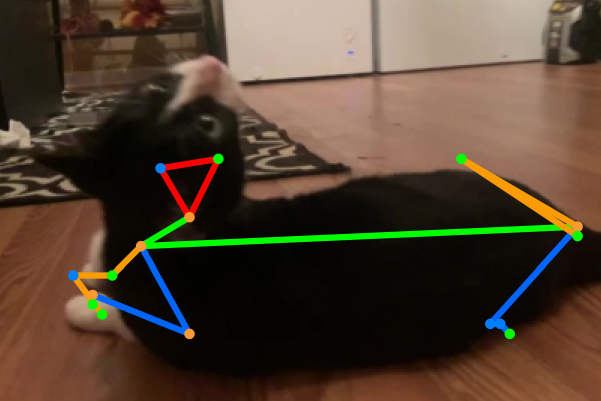}}\\ 
\end{tabular}
\caption{Example pose predictions: (a-b) CrowdPose predictions using the COCO skeleton; (c--d) COP3D predictions using the AP-10K skeleton.
}
\label{fig:supp_crowdpose}
\end{figure*}

\section{Additional results}
\label{s:additionalresults}
\subsection{Quantitative results}
\Cref{table:supp_coco_test} presents the pose estimation results on the COCO test-dev set. Following previous works~\cite{xiao2018simple,sun2019deep,li2022simcc}, we cropped the input images based on detected bounding boxes. Our method outperforms the baseline ViTPose++ by 0.2 AP with the \emph{Base} backbone and by 0.1 AP with the \emph{Huge} backbone.

We also performed a downstream evaluation on unseen data using the OCHuman~\cite{zhang2019pose2seg} dataset, which comprises 2,500 validation images and 2,231 test images, with no available training set. The results are presented in~\cref{table:supp_ochuman}. Since OCHuman follows the COCO keypoint format, we used COCO-trained models for evaluation. Here, `(occ)' denotes the evaluation of the occluded keypoints, following the protocol of~\cite{Geng23PCT}. Our method outperforms the baseline ViTPose++, particularly on the (occ) subsets, demonstrating its robustness to occlusion.

Pose estimation lacks a large, generic, high-quality dataset, which is a key motivation for our multi-dataset training (MDT) approach. As shown in \cref{table:supp_scratch}, our method outperforms models trained from scratch or transferred from the \emph{largest} single dataset (AIC) in domain transfer scenarios for pose estimation. Furthermore, compared to existing MDT problems (\eg, classification and detection), skeletal heterogeneity in pose estimation presents a unique challenge, making na\"ive dataset merging or multi-head supervision ineffective. \Cref{table:supp_label_agg} demonstrate this: Our method significantly outperforms conventional label merging approaches when applied to pose datasets such as UniDet~\cite{zhou2022simple}.

\subsection{Qualitative results}
\Cref{fig:human_vis_suppl} presents additional human pose estimation examples. In the first two rows, ViTPose++ struggles with accurate leg estimation and exhibits inconsistent keypoint predictions across different dataset skeletons. In contrast, our method accurately estimates the legs and maintains consistency across varying skeletons. In the bottom two rows, ViTPose++ incorrectly predicts the right and left foot at the same location, whereas our method correctly estimates the legs, except for AIC.

\Cref{fig:animal_vis_suppl_1} provides additional comparisons on the AP-10K animal dataset. For cattle (first two columns), ViTPose++ mislocalizes the left front leg in the APT-36K skeleton, while our method correctly identifies it. Similarly, in the last two columns (kangaroo), ViTPose++ confuses the right front leg with the left, an error our method successfully avoids.

\Cref{fig:interhand_suppl} and \Cref{fig:3dpw_suppl} provide additional pose estimation examples of our algorithm on the InterHand2.6M and 3DPW datasets, respectively. Our method demonstrates strong generalization across hand and human shapes, even under various self-occlusion and external occlusion scenarios.

We visualize the prototypes constructed by our algorithm in \cref{fig:emb_vis_suppl}, using those trained with the ViT-B backbone. The InterHand and 3DPW prototypes are separately trained during the domain transfer process, while others are jointly learned during MDT. The prototypes effectively capture the diversity of representations within the embedding space and successfully align similar keypoints across different datasets without compromising localization performance. 

For example, in the upper red box in the figure, the COCO nose joint and the 3DPW jaw joint prototypes are closely aligned. Similarly, the lower red box contains a COCO left hip joint prototype and a 3DPW pelvis joint. As validated by domain transfer on InterHand and 3DPW, our learned embeddings effectively incorporate new skeletons without retraining of the embedding module.

\subsection{Failure cases}
In~\cref{fig:supp_crowdpose}, we provide failure cases caused by unseen skeletons and poses. In (a--b), we show our COCO skeleton predictions on CrowdPose data. Since CrowdPose has a different skeletal structure than the training datasets, the predictions do not fully conform to its intended format, although the pose is reasonably well estimated in (a). Under strong occlusion (b), our method may also struggle to predict accurate skeletons, as seen in the misplacement of the estimated left foot at the location of the right foot. 

Similarly, (c--d) show our AP-10K skeleton predictions on a COP3D~\cite{sinha2023common} example, where the pose deviates significantly from those observed during training (\eg the cat’s head is tilted back, looking upward). In (d), our method fails to localize the cat’s eyes due to this challenging, unseen pose. Incorporating temporal information could improve robustness in such cases.


\end{document}